\RequirePackage[svgnames]{xcolor}

\documentclass[11pt,letterpaper]{mystyle}
\usepackage[all]{hypcap}
\usepackage[svgnames]{xcolor}
\usepackage[comma,authoryear,compress]{natbib}
\usepackage{hyperref}
\usepackage{amsmath}
\usepackage{etoolbox}

\hypersetup{
    colorlinks = true,
    citecolor = {YaleBlue},
}

\usepackage{wrapfig}

\usepackage{algorithm}
\usepackage{algorithmicx}
\usepackage{algpseudocode}
\usepackage{microtype}
\usepackage{booktabs} %
\usepackage{float}
\usepackage{bigstrut}

\usepackage{comment}
\usepackage{amsmath}
\usepackage{amssymb}
\usepackage{mathtools}
\usepackage{amsthm}
\usepackage{mathrsfs}
\usepackage{nicefrac}
\usepackage{dsfont}
\usepackage{enumitem}
\usepackage{subcaption}
\usepackage{cleveref}
\usepackage{bxcoloremoji}
\usepackage{float}
\usepackage{algpseudocode}

\usepackage[utf8]{inputenc} %
\usepackage[T1]{fontenc}    %
\usepackage{url}            %
\usepackage{booktabs}       %
\usepackage{amsfonts}       %
\usepackage{nicefrac}       %
\usepackage{microtype}      %
\usepackage{graphicx}
\usepackage{amssymb}
\usepackage{fdsymbol}
\usepackage{wrapfig}
\usepackage{lipsum}
\usepackage{enumitem}
\usepackage{stackengine}
\usepackage[font=small,labelfont=bf]{caption}
\usepackage{color}
\usepackage{adjustbox}

\usepackage{rotating}
\usepackage{makecell}

\usepackage{multirow}

\definecolor{BlueBG}{rgb}{0,0.46,0.71}

\newcommand{\x}{\mathbf{x}}

\definecolor{blanchedalmond}{rgb}{1.0, 0.92, 0.8}
\definecolor{carmine}{rgb}{0.59, 0.0, 0.09}
\definecolor{lightblue}{rgb}{0.22,0.45,0.70}%

\newtheorem{theorem}{Theorem}[section]

\newtheorem{definition}[theorem]{Definition}

\renewcommand{\mathbf}{\boldsymbol}

\makeatletter
\def\Ddots{\mathinner{\mkern1mu\raise\p@
\vbox{\kern7\p@\hbox{.}}\mkern2mu
\raise4\p@\hbox{.}\mkern2mu\raise7\p@\hbox{.}\mkern1mu}}
\makeatother

\definecolor{amaranth}{rgb}{0.9, 0.17, 0.31}
\definecolor{antiquebrass}{rgb}{0.8, 0.58, 0.46}
\definecolor{antiquefuchsia}{rgb}{0.57, 0.36, 0.51}
\definecolor{chromeyellow}{rgb}{0.31, 0.47, 0.26}

\usepackage{url}
\usepackage{xcolor}
\usepackage{pifont}
\usepackage{soul}

\usepackage{amsmath}

\usepackage[all]{hypcap}

\usepackage{algpseudocode}
\usepackage{authblk}   
\usepackage{amssymb}      
\usepackage{bbm}          
\usepackage{optidef}

\usepackage[utf8]{inputenc}
\usepackage[T1]{fontenc}

\usepackage{mathpazo}
\usepackage[svgnames]{xcolor}

\usepackage{lipsum}

\usepackage{pifont} 
\usepackage{xspace}
\newcommand{\ourmethod}{{\fontfamily{lmtt}\selectfont \textbf{SkillOrchestra}}\xspace}
\usepackage{ulem}
\usepackage{amsmath}
\usepackage{enumerate}
\usepackage{enumitem}
\usepackage{tcolorbox}
\usepackage{tikz}
\usetikzlibrary{positioning, arrows.meta}

\usepackage[ruled,algo2e,vlined]{algorithm2e}
\SetKwInOut{Input}{Input}
\SetKwInOut{Output}{Output}
\SetKwComment{Comment}{$\triangleright$\ }{}
\DontPrintSemicolon

\usepackage{multirow}
\usepackage{makecell}
\usepackage{arydshln}

\definecolor{myred}{HTML}{EF8580}
\definecolor{myyellow}{HTML}{FDCC91}
\definecolor{myblue}{HTML}{8BC0E7}

\definecolor{mygrey}{RGB}{120,120,120}
\definecolor{Periwinkle}{RGB}{120,140,255}

\newcommand{\tagpill}[2]{%
  {\colorbox{#1!12}{\textcolor{#1}{\strut\textbf{\footnotesize\ #2\ }}}}%
}

\newcommand{\statuspill}[2]{%
  {\colorbox{#1!10}{\textcolor{#1}{\strut\footnotesize\textbf{#2}}}}%
}

\newcommand{\best}[1]{\textbf{#1}}
\newcommand{\second}[1]{\underline{#1}}

\newcommand{\cmark}{\ding{52}} % check mark
\newcommand{\xmark}{$\circ$}   % empty circle

\definecolor{darksalmon}{rgb}{0.91, 0.59, 0.48}
\definecolor{mygrey}{RGB}{120,120,120}
\definecolor{Periwinkle}{RGB}{120,140,255}
\definecolor{TagCyan}{RGB}{0,170,200}
\definecolor{TagBlue}{RGB}{25,70,160}
\definecolor{TagGreen}{RGB}{30,140,80}
\definecolor{TagOrange}{RGB}{210,120,0}

\usepackage{booktabs}
\usepackage{bbding}

\usepackage{siunitx}
\usepackage{caption}
\usepackage{subcaption}
\usepackage{listings}
\definecolor{mygrey}{gray}{0.4}

\usepackage{listings}
\usepackage{fvextra}   % better verbatim
\tcbuselibrary{skins,breakable}

\usepackage{fontawesome5}

\title{SkillOrchestra: Learning to Route Agents via Skill Transfer}

\runningtitle{SkillOrchestra: Learning to Route Agents via Skill Transfer}

\author[1]{Jiayu Wang}
\author[2]{Yifei Ming}
\author[2]{Zixuan Ke}
\author[2]{Shafiq Joty}
\author[1]{Aws Albarghouthi}
\author[1]{Frederic Sala}

\affil[1]{University of Wisconsin-Madison}
\affil[2]{Salesforce AI Research}

\correspondingauthor{Jiayu Wang: milawang@cs.wisc.edu}

\begin{document}

\begin{abstract}
{\centering\section*{Abstract}}
Compound AI systems promise capabilities beyond those of individual models, yet their success depends critically on effective orchestration.
Existing routing approaches face two limitations: (1) input-level routers make coarse query-level decisions that ignore evolving task requirements; (2) RL-trained orchestrators are expensive to adapt and often suffer from \textit{routing collapse}, repeatedly invoking one strong but costly option in multi-turn scenarios.
We introduce \ourmethod, a framework for \textbf{skill-aware orchestration}.
Instead of directly learning a routing policy end-to-end, \ourmethod learns fine-grained skills from execution experience and models agent-specific competence and cost under those skills.
At deployment, the orchestrator infers the skill demands of the current interaction and selects agents that best satisfy them under an explicit performance-cost trade-off.
Extensive experiments across ten benchmarks demonstrate that \ourmethod outperforms SoTA RL-based orchestrators by up to 22.5\% with 700$\times$ and 300$\times$ learning cost reduction compared to Router-R1 and ToolOrchestra, respectively.  These results show that explicit skill modeling enables scalable, interpretable, and sample-efficient orchestration, offering a principled alternative to data-intensive RL-based approaches. The code is available at: \url{https://github.com/jiayuww/SkillOrchestra}.
\end{abstract}

\maketitle

\section{Introduction}
\label{sec:intro}

\begin{figure}[h!]
    \centering
    \includegraphics[width=1\linewidth]{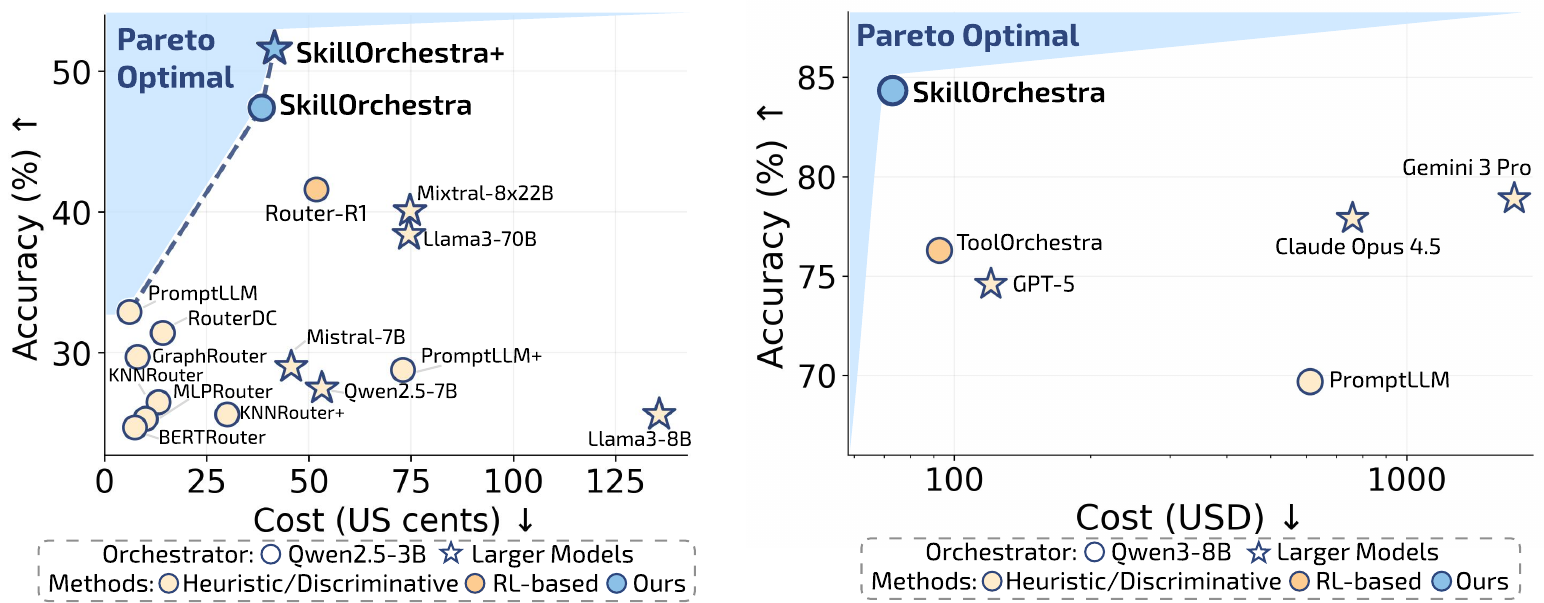}
    \caption{Performance-cost tradoffs in multi-turn model routing (left) and agent orchestration (right). SkillOrchestra and SkillOrchestra+ lie on the Pareto frontier, with higher accuracy at lower cost than all baselines.}
    \label{fig:pareto_overview}
\end{figure}

\begin{figure}[th!]
    \centering
    \includegraphics[width=1\linewidth]{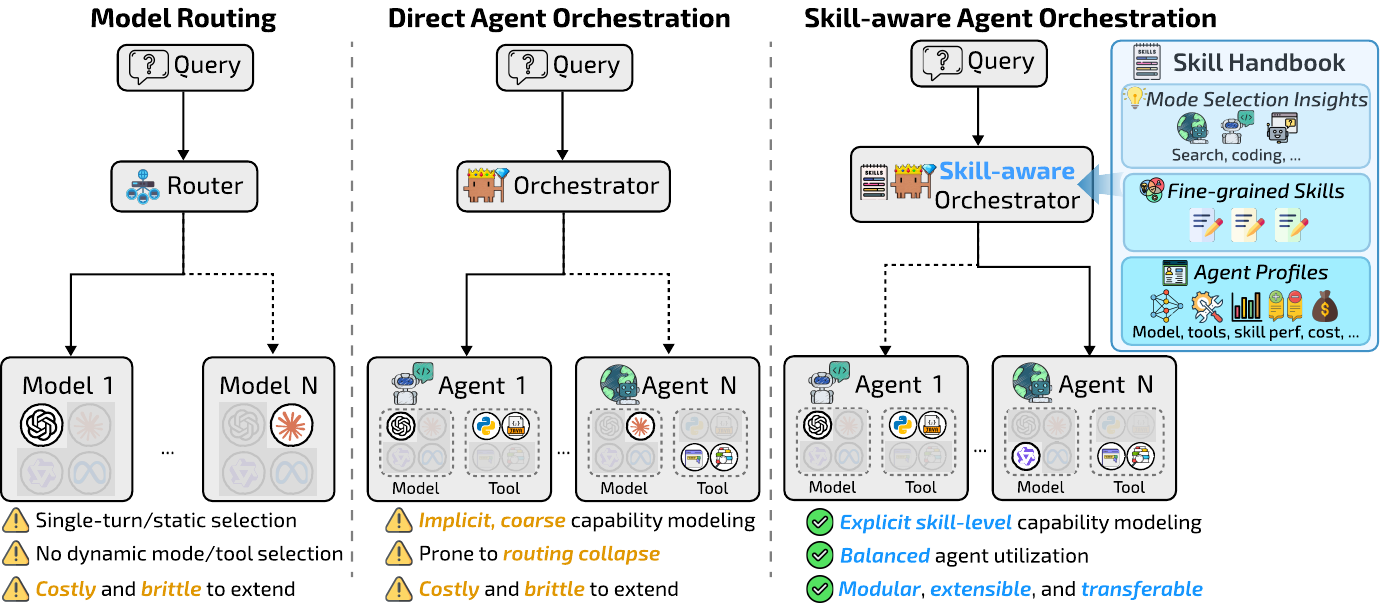}
    \caption{Comparison of model routing and agent orchestration approaches.
\textbf{(Left)} Model routing performs static, query-level model selection without dynamic mode or tool reasoning.
\textbf{(Middle)} Direct agent orchestration learns routing end-to-end with implicit capability modeling and is prone to routing collapse.  
\textbf{(Right)} Skill-aware agent orchestration leverages a reusable Skill Handbook with explicit skill-level capability modeling, enabling balanced agent utilization and extensibility.}
    \label{fig:route_orch_skill_comp}
\end{figure}

Modern AI systems are increasingly built as compound agents that coordinate multiple large language models (LLMs) and tools to solve complex, multi-step tasks such as deep research~\citep{gemini_deep_research,openai_deep_research} and scientific discovery~\citep{gottweis2025aicoscientist}. Instead of relying on a single model, these systems interleave operations such as web search, code execution, and answer synthesis, dynamically invoking models with different strengths and costs~\citep{ke2025a}. In this setting, \textit{orchestration}, the process deciding what capability is required at each interaction state and which model--tool combination to invoke, is central to both performance and efficiency.

A common form of orchestration is model routing, where a controller selects a model from a model pool~\citep{chen2024frugalgpt,hu2024routerbench,ong2025routellm}. However, existing routing methods are often ill-suited to modern agentic workloads. Most routers make single-shot, query-level decisions, assuming one model suffices for the entire task. This assumption breaks down in multi-turn interactions, where different states require distinct capabilities.
Agentic workflows often interleave operational modes (e.g., web search and coding), each demanding different skills. Routing should therefore operate at the level of fine-grained capability requirements conditioned on the current interaction state, rather than treating the entire query as a single decision unit (Figure~\ref{fig:route_orch_skill_comp}, left).
Recent RL-based orchestration methods~\citep{zhang2025routerr,toolorchestra} address this by learning sequential routing policies with LLMs. 
While more flexible, these approaches introduce new challenges: expensive training, limited adaptability to evolving model and tool pools, and a tendency toward what we term \textit{routing collapse}: the degeneration of the orchestration policy into repeatedly selecting a single option at one or more decision levels (e.g., agent type or backbone model), despite the availability of alternatives with better accuracy-cost trade-offs (Figure~\ref{fig:route_orch_skill_comp}, middle).

To address these limitations, we introduce \textbf{SkillOrchestra}, a framework for \textit{skill-aware orchestration}. 
Rather than directly optimizing a routing policy end-to-end, SkillOrchestra learns a reusable Skill Handbook from execution experience. 
The handbook encodes (i) mode-level execution insights that guide what operation should be performed at each interaction state, (ii) fine-grained skills that characterize capability requirements within each mode,
and (iii) agent profiles that summarize skill-conditioned performance, cost characteristics, and practical usage insights.
At deployment, the orchestrator first selects the appropriate operational mode conditioned on the current state, then chooses the agent that best satisfies the required skills under an explicit performance-cost trade-off (Figure~\ref{fig:route_orch_skill_comp}, right).

This skill-centric perspective brings three systemic advantages.
First, it enables state-conditioned, fine-grained orchestration, allowing different models to specialize across capabilities. 
Second, it promotes stable and balanced routing behavior, mitigating routing collapse seen in RL-tuned orchestrator. 
Third, it produces transferable orchestration knowledge: the learned Skill Handbook can be reused across different orchestrator backbones and updated model pools, decoupling orchestration knowledge from router parameters.

We evaluate SkillOrchestra in both multi-turn model routing and full agent orchestration settings. 
As shown in Figure~\ref{fig:pareto_overview}, SkillOrchestra and SkillOrchestra+ lie on the Pareto frontier, achieving higher accuracy at lower cost than all baselines. 
Across ten diverse benchmarks, SkillOrchestra consistently outperforms heuristic, discriminative, and RL-based approaches. 
For example, SkillOrchestra outperforms SoTA RL-trained orchestrators, achieving up to 22.5\% absolute improvement, with 700$\times$ and 300$\times$ cost reduction compared to Router-R1~\citep{zhang2025routerr} and ToolOrchestra~\citep{toolorchestra}, respectively. 
Moreover, it exhibits more balanced routing patterns and transfers effectively across orchestrator models without retraining. We summarize our contributions as follows:

\textbf{\ding{182} Skill-aware orchestration.} We propose SkillOrchestra, a new paradigm that structures orchestration decisions around explicit capability abstractions and agent profiles, enabling state-conditioned, performance-cost-aware orchestration.

\textbf{\ding{183} Skill Handbook learning.} We introduce a data-efficient framework to discover and refine reusable skills and execution insights from agent traces, while estimating skill-conditioned agent performance and cost.

\textbf{\ding{184} Granularity-aware skill handbook selection.} We show that optimal skill granularity depends on orchestrator capacity, and develop a validation strategy to select orchestrator-specific handbooks that balance expressiveness and decision reliability under performance-cost trade-offs.

\textbf{\ding{185} Empirical gains and transferability.} Extensive experiments across ten benchmarks demonstrate improved accuracy, efficiency, and routing stability over strong RL-tuned baselines, alleviating routing collapse and transferring across orchestrator backbones without retraining.

\section{Related Works}\label{sec:related}
\textbf{Model Routing.}
Model routing aims to select the most appropriate model from a pool to balance performance and inference cost. Early approaches rely on heuristic or cascade strategies~\citep{chen2024frugalgpt} that escalate queries based on predicted difficulty or budget constraints~\citep{ding2024hybrid,vsakota2024fly}. Prior approaches are largely heuristic or discriminative, learning static mappings from query features to model choice~\citep{blender} or relying on cascades~\citep{chen2024frugalgpt} and difficulty estimation.
Instead, a large body of work learns discriminative query-model matching, using similarity-based methods~\citep{hu2024routerbench,ong2025routellm}, neural classifiers or ensembles~\citep{blender,zooter}, and graph-based formulations~\citep{feng2025graphrouter} to predict which model should answer a query~\citep{chen2024routerdc,stripelis-etal-2024-tensoropera}.
Despite their effectiveness, routing decisions for these approaches are typically made once per query using input-level features only, without modeling how model competence differs across intermediate stages. As a result, they struggle to support fine-grained, multi-step orchestration.

\textbf{RL-based Routing and Orchestration.} To enable multi-step decisions, recent work formulates routing as a sequential decision process and trains an LLM-based router using reinforcement learning~\citep{schulman2017proximal,shao2024deepseekmath}. Systems such as Router-R1~\citep{zhang2025routerr} and ToolOrchestra~\citep{toolorchestra} that interleave reasoning and routing, optimizing performance-cost trade-offs via trajectory-level rewards. 
While more flexible than single-shot routers, RL-based approaches introduce new challenges such as high training cost, poor adaptability to new model pools or tasks, and policy routing collapse, where the router converges to repeatedly invoking a single strong but expensive model.
In contrast, we introduce \textbf{\textit{skill}} as an intermediate abstraction and construct a reusable Skill Handbook that captures mode-conditioned competence patterns.
This design enables data-efficient, transferable, and more balanced orchestration without end-to-end RL training.

\begin{figure}[t!]
    \centering
    \includegraphics[width=1\linewidth]{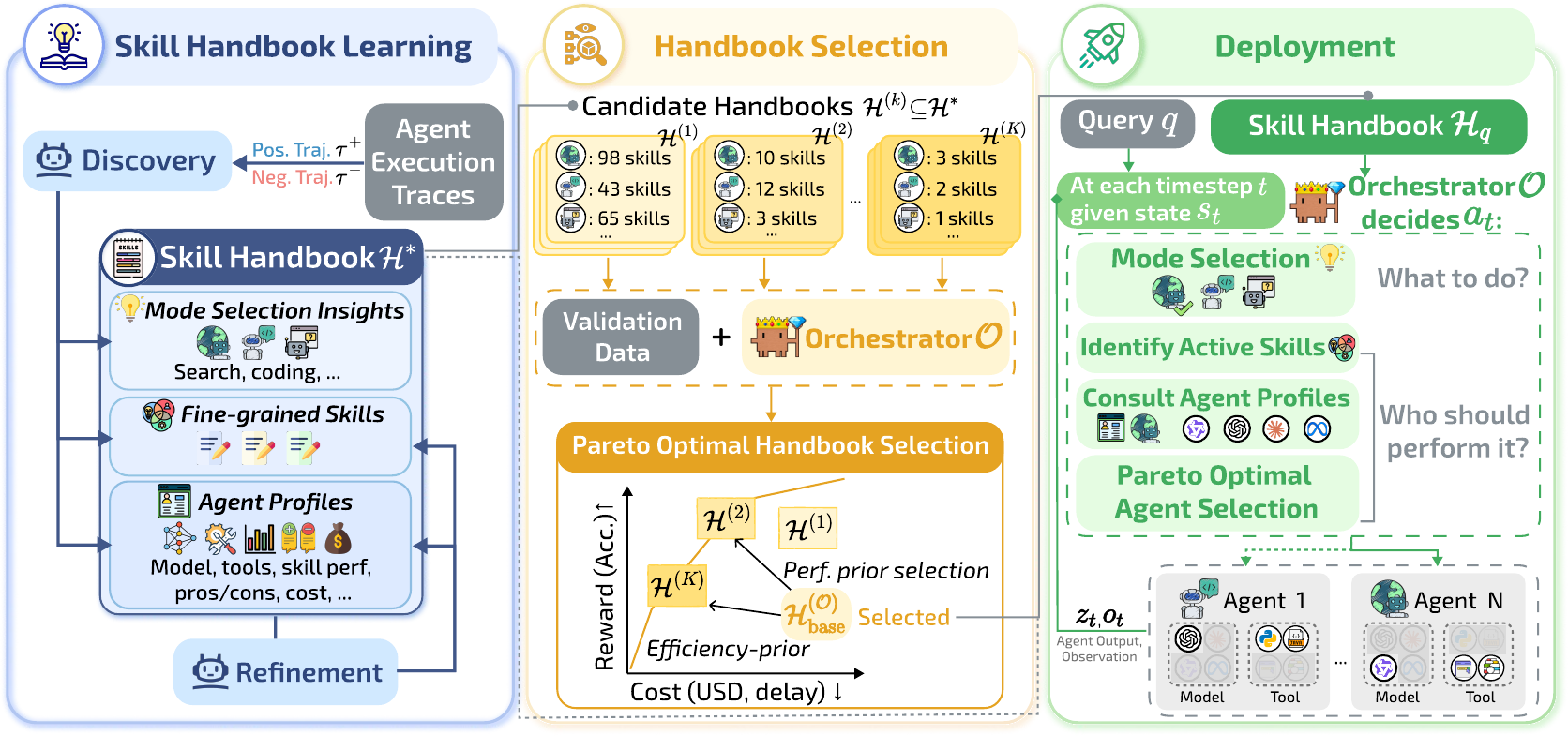}
    \caption{Overview of SkillOrchestra. \textbf{(Left)} A global Skill Handbook is constructed by discovering and refining reusable skills and execution-level insights from agent traces, while jointly estimating each agent’s skill competence and associated cost. 
\textbf{(Middle)} An orchestrator-specific handbook is selected via Pareto validation to achieve a principled trade-off between performance and cost. 
\textbf{(Right)} At deployment, the orchestrator performs mode-aware and skill-grounded agent selection using the selected handbook.}
    \label{fig:skillorchestra_overview}
\end{figure}

\section{Preliminaries}\label{sec:prelim}
\paragraph{Agent Orchestration.}
We consider an agentic task environment where a user instruction $q \in \mathcal{Q}$ initiates a multi-step reasoning process. The system consists of the following components:

[\ding{81}] The Orchestrator ($\mathcal{O}$): A central controller responsible for high-level planning and resource allocation.

[\ding{81}] Operational Modes ($\Psi$). A set of abstract action modes $\Psi = \{\texttt{search}, \texttt{code}, \dots\}$ defined at the capability level. At each turn, the orchestrator chooses a mode $\psi \in \Psi$ that specifies the type of operation required (e.g., retrieving external information, or code execution).

[\ding{81}] Model Pool ($\mathcal{M}$).
A set of candidate foundation models $\mathcal{M} = \{m_1, \dots, m_{K_M}\}$, which may include general-purpose and specialized LLMs (e.g., GPT-5, Claude, Qwen-3, or domain-specific coder and math models).

[\ding{81}] Tool Pool ($\mathcal{T}$).
A set of executable tools
$
\mathcal{T} = \{ t_1, \dots, t_{K_T} \},
$
such as web search engines (e.g., Google Search, Tavily Search), code execution environments (e.g., Python), database retrieval systems, or other external APIs.

[\ding{81}] Agent Instantiation.
An \textit{agent} is defined as a pair
\[
A = (m, \mathcal{T}_A), \quad m \in \mathcal{M}, \; \mathcal{T}_A \subseteq \mathcal{T},
\]
where $m$ is the backbone model and $\mathcal{T}_A$ is the subset of tools accessible during execution.

Each operational mode $\psi \in \Psi$ restricts the allowable tools, inducing a set of valid agents
\[
\mathcal{A}_\psi = \{(m, \mathcal{T}_A) \mid m \in \mathcal{M},\; \mathcal{T}_A \subseteq \mathcal{T}_\psi \},
\]
where $\mathcal{T}_\psi \subseteq \mathcal{T}$ denotes the tools relevant to mode $\psi$.

\textbf{Task Execution Workflow.}
Given query $q$, the system evolves over turns $t=0,\dots,T$.
Let $s_t$ denote the system state at turn $t$, which consists of the original query and the accumulated interaction history up to that point. 
At each turn, the orchestrator selects a mode $\psi_t \in \Psi$ (what to do) and an agent $A_t \in \mathcal{A}_{\psi_t}$ (who executes it), forming the action $a_t = (\psi_t, A_t)$. 
The selected agent produces an execution trace $z_t$ (e.g., search results or generated code), after which the environment returns an observation $o_t$ (e.g., tool outputs or execution results), leading to the next state $s_{t+1}$. 
This interaction induces a trajectory
$\tau = (s_0, a_0, z_0, o_0, s_1, a_1, z_1, o_1, \dots, s_T)$.
An example multi-step workflow is illustrated in Fig.~\ref{fig:skillorchestra_overview} (right).

\textbf{Problem Formulation.} 
The orchestrator aims to learn a policy $\pi$ that optimizes performance-cost tradeoffs over trajectories. 
Formally, we seek to maximize the expected reward $R(\tau)$ and minimizing the cumulative execution cost:
\begin{equation*}
\max_\pi J(\tau) = \mathbb{E}_{\tau \sim \pi} \left[ R(\tau) - \lambda \sum_{t=0}^T C(A_t, z_t) \right],
\end{equation*}
where $C(A_t, z_t)$ denotes the cost incurred by the selected agent $A_t$ when producing trace $z_t$ (e.g., token usage and/or latency), and $\lambda$ is a tradeoff hyperparameter.
We factorize the policy as
\begin{equation*}
\pi(a_t \mid s_t) = \pi_{\text{mode}}(\psi_t \mid s_t)\cdot \pi_{\text{route}}(A_t \mid s_t, \psi_t),
\end{equation*}
where $\pi_{\text{mode}}$ determines the next operational mode (e.g., \textit{Search} vs. \textit{Coding}), and $\pi_{\text{route}}$ selects the optimal agent $A_t$ conditioned on the current state and mode.

Under this formulation, traditional model routing~\citep{hu2024routerbench,chen2024routerdc} can be viewed as a special case with a single timestep $T=0$, a single operational mode $\Psi={\textit{answer}}$, and no external tools. The objective reduces to $\max_{\pi_{\text{route}}} \mathbb{E}_{A \sim \pi_{\text{route}}(\cdot \mid q)}
\left[R(A, q) - \lambda C(A, z)\right]$, where the state $s_0=q$ is the user query and routing consists of choosing one model to generate the final answer in a single step.

Prior work typically instantiates this optimization via RL such as GRPO~\citep{shao2024deepseekmath} by directly finetuning the orchestrator parameters $\theta$ toward the optimal policy~\citep{toolorchestra}. In contrast, \ourmethod reframes orchestration as a problem of \textbf{skill acquisition} rather than parameter adaptation. Instead of updating $\theta$, we learn a \textbf{Skill Handbook} $\mathcal{H}$, a reusable experience base that captures (i) mode-level execution insights about what operation to perform at a given interaction state, (ii) fine-grained skills that characterize capability requirements within each mode, and (iii) agent profiles that summarize competence and cost under those skills (e.g., \textit{high-precision arithmetic}, \textit{symbolic logic coding}).
Under this view, the optimization shifts from learning a routing policy to identifying the optimal handbook structure:
\begin{equation*}
\mathcal{H}^* = \operatorname*{argmax}_{\mathcal{H}} \mathbb{E}_{\tau \sim \pi (\cdot \mid \mathcal{H})} \left[ J(\tau) \right].
\end{equation*}
By optimizing the Skill Handbook $\mathcal{H}$, we explicitly align abstract task demands with concrete agent capabilities, enabling the orchestrator to reason over the competence landscape of the agent pool even without costly end-to-end RL finetuning.

\section{SkillOrchestra}\label{sec:method}
\ourmethod reframes orchestration as skill-grounded decision making rather than direct policy optimization. 
Instead of learning a monolithic routing policy, we learn a structured Skill Handbook that captures reusable execution knowledge.
During training, the handbook is incrementally constructed and refined from execution traces, including skills, agent profiles, and execution insights. 
At test time, the orchestrator consults a selected subset of this handbook to guide mode selection and agent routing.

\begin{definition}[Skill]
\label{def:skill}
A \textit{skill} is a reusable capability abstraction that specifies the type of competence required to perform a task under an operational mode $\psi$. 
Skills form an intermediate layer between high-level modes (e.g., $\texttt{search}$, $\texttt{code}$) and individual agents, enabling the system to decouple capability requirements from agent identity.

Formally, a skill $\sigma$ is represented as
\[
\sigma \triangleq \langle \mathcal{D}, \mathcal{I} \rangle,
\]
where $\mathcal{D}$ is a natural-language description of the capability, and 
$\mathcal{I}$ denotes contextual indicators (e.g., keywords, structural patterns, or exemplar queries) that signal when the skill is applicable.
\end{definition}

\begin{definition}[Agent Profile]
\label{def:agent_profile}
An \textit{agent profile} summarizes an agent’s mode-conditioned competence, cost, and routing characteristics for skill-aware orchestration. 
For agent $A$ under operational mode $\psi$, the profile is defined as
\begin{equation*}
\mathcal{P}_{A,\psi}
=
\left(
\{\phi_{A,\sigma}\}_{\sigma \in \Sigma_{\psi}},
\hat{C}_A(\psi),
\mathcal{R}_{A,\psi},
\Gamma_A
\right),
\end{equation*}
where
$\phi_{A,\sigma}$ denotes the estimated success probability of agent $A$ on skill $\sigma$,
$\hat{C}_A(\psi)$ is the estimated execution cost (e.g., latency, token usage) under mode $\psi$,
$\mathcal{R}_{A,\psi}$ encodes mode-conditioned routing signals such as usage constraints or systematic failures,
$\Gamma_A$ provides a high-level summary of the agent’s strengths and weaknesses.
\end{definition}

\subsection{Agent Orchestration via Skill Handbook}\label{sec:inference}
We now describe runtime orchestration using the Skill Handbook (Fig.~\ref{fig:skillorchestra_overview}, right).

\begin{figure}[t!]
    \centering
    \includegraphics[width=1\linewidth]{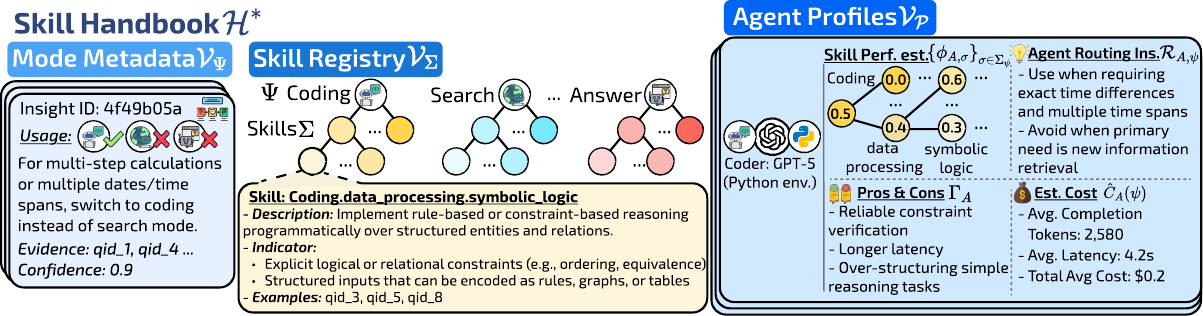}
    \caption{Example instantiation of a learned Skill Handbook. 
The handbook decouples capability requirements from agent identity through three components: 
(left) mode-level routing insights, 
(middle) a hierarchical registry of reusable skills, and 
(right) agent profiles encoding skill-specific competence estimates and execution cost statistics. }
    \label{fig:skill_handbook}
\end{figure}

\textbf{Skill Handbook.}
The Skill Handbook $\mathcal{H}$ organizes reusable orchestration knowledge at three levels:
(i) mode-level execution insights that guide what operation to perform under different interaction states,
(ii) a registry of fine-grained skills that capture capability requirements within each mode,
and (iii) agent profiles that model skill-conditioned competence, routing signals, and execution cost. 
It can be viewed as a graph $\mathcal{G}_{\mathcal{H}} = (\mathcal{V}, \mathcal{E})$, $\mathcal{V} = \mathcal{V}_{\Psi} \cup \mathcal{V}_{\Sigma} \cup \mathcal{V}_{\mathcal{P}}$ stores mode selection insights, skills, and agent profiles.
The edge structure encodes associations between operational modes and relevant skills.

\textbf{Example (Skill Handbook Instantiation).}
Figure~\ref{fig:skill_handbook} shows a concrete instantiation of $\mathcal{H}$.
For example, under mode $\psi=\texttt{code}$, the handbook stores mode-level metadata (left) capturing when to code.
The skill registry (middle) may include a high-level skill \texttt{data\_processing}, which further specializes into subskills such as \texttt{symbolic\_logic}.
Each agent is associated with a profile (right) providing competence estimates over these skills, mode-conditioned routing signals, and execution cost statistics.
Together, these components enable structured, skill-grounded agent selection.

[\ding{81}] \textbf{Mode-level metadata $\mathcal{V}_\Psi$.}
For each operational mode $\psi \in \Psi$, the handbook stores mode-level routing insights $\mathcal{R}_\psi$ learned from execution traces, guiding high-level transitions (e.g., when to switch from \textit{Search} to \textit{Code}).

[\ding{81}] \textbf{Skill registry $\mathcal{V}_\Sigma$.}
The handbook maintains a registry of skills
(Definition~\ref{def:skill}),
each representing a task-conditioned capability that may be required during execution.

[\ding{81}] \textbf{Agent profiles $\mathcal{V}_{\mathcal{P}}$.}
Each agent $A$ is associated with an agent profile (Definition~\ref{def:agent_profile}), which stores agent-specific performance estimates over skills, routing insights of this agent, and cost characteristics.
Agent profiles are queried during routing but are not indexed by graph edges.

[\ding{81}] \textbf{Mode--skill index $\mathcal{E}$.} 
The graph structure induces a mapping
$M : \Psi \rightarrow 2^{\Sigma}$,
where $\Sigma_\psi := M(\psi)$ denotes the set of skills associated with operational mode $\psi$.
This index restricts routing decisions to mode-consistent skills without searching over the full skill space at runtime.

\textbf{Orchestration with Skill Handbook.}
At inference time, the orchestrator interacts with the Skill Handbook in a task-conditioned manner.
Given a user query $q$, the system follows a \textit{retrieval--execution} cycle.

\textbf{Step 1: Handbook Selection.} 
The effectiveness of a handbook depends on how well its structural granularity aligns with the reasoning capacity of the target orchestrator.
Although the learned handbook $\mathcal{H}^*$ may contain fine-grained skills and detailed routing insights derived from prior experience, not all such structure is equally beneficial for every orchestrator.

Fine-grained skill decompositions require accurate inference of which subskill is active in the current interaction state.
While a strong orchestrator may reliably distinguish between subskills such as 
\texttt{symbolic\_logic} and \texttt{numerical\_approximation} under \texttt{code} mode,
a lower-capacity orchestrator may misidentify the active skill, introducing routing bias and degrading end-to-end performance.
For example, in a coding query requiring logical constraint verification, activating \texttt{numerical\_approximation} instead of \texttt{symbolic\_logic} may route to an agent specialized in numeric computation but suboptimal for symbolic reasoning.
Operating at a coarser granularity (e.g., using a broader skill such as \texttt{data\_processing}) reduces sensitivity to such misidentification and yields more stable routing decisions.

Starting from the learned handbook $\mathcal{H}^*$ (Section~\ref{sec:learning}),
we therefore select an orchestrator-specific subset
$\mathcal{H}_{\text{base}}^{(\mathcal{O})}$ for orchestrator $\mathcal{O}$ via
Pareto-optimal validation (Section~\ref{sec:selection}).
This selection determines which skills, agent profiles, and routing metadata are
retained, as well as their effective granularity, so as to maximize end-to-end
performance given target orchestrator under a given cost budget.
Formally, $\mathcal{H}_{\text{base}}^{(\mathcal{O})}$ is an induced subgraph of
$\mathcal{H}^*$:
$
\mathcal{H}_{\text{base}}^{(\mathcal{O})}
=
\big(
\mathcal{V}_{\Psi}^{\text{base}}
\;\cup\;
\mathcal{V}_{\Sigma}^{\text{base}}
\;\cup\;
\mathcal{V}_{\mathcal{P}}^{\text{base}},
\mathcal{E}^{\text{base}}
\big),
$
where
$\mathcal{V}_{\Psi}^{\text{base}}$ contains mode-level routing metadata useful for the orchestrator to select operational modes,
$\mathcal{V}_{\Sigma}^{\text{base}}$ contains the skills retained for those modes at the selected granularity,
and $\mathcal{V}_{\mathcal{P}}^{\text{base}}$ contains the corresponding agent profiles.
The edge set $\mathcal{E}^{\text{base}} \subseteq \mathcal{E}$ restricts mode--skill associations to the retained nodes.
All node attributes, including routing insights, performance estimates, and cost statistics, are inherited from $\mathcal{H}^*$.

At inference time, the orchestrator retrieves $\mathcal{H}_{\text{base}}^{(\mathcal{O})}$.
Optionally, the retrieval operator may augment $\mathcal{H}_{\text{base}}^{(\mathcal{O})}$
with additional skills whose semantic similarity to the query exceeds a
threshold, yielding the final handbook $\mathcal{H}_q$ used for query $q$ and orchestrator $\mathcal{O}$:
$
% \begin{equation}
\mathcal{H}_q
=
\mathcal{H}_{\text{base}}^{(\mathcal{O})}
\;\cup\;
\bigcup_{\sigma \in \mathcal{N}_k(q)}
\left(
\{\sigma\}
\;\cup\;
\{\mathcal{P}_{A,\psi} \mid A \in \mathcal{A}_\psi\}
\right),
% \end{equation}
$
where $\mathcal{N}_k(q)$ is the $k$ nearest skills in the embedding space.

\textbf{Step 2: Skill-Grounded Agent Routing.}
Guided by the retrieved handbook $\mathcal{H}_q$, the orchestrator performs skill-grounded routing through an iterative decision process. An illustration can be found in Figure~\ref{fig:skillorchestra_overview} (right).
At each time step $t$, it decides:

\ding{70} \textbf{Mode Selection.}
The mode policy $\pi_{\text{mode}}$ selects the current operational mode $\psi_t$ based on the interaction state $s_t$ and the mode-level routing metadata stored in the handbook:
% \begin{equation}
$
\psi_t \sim \pi_{\text{mode}}(\psi \mid s_t; \mathcal{R}_{\psi}).
$
% \end{equation}
This decision determines the operational mode to execute next (e.g., \textit{Search},
\textit{Code}).

\ding{70} \textbf{Competence-Aware Agent Routing.}
Conditioned on the selected mode $\psi_t$, the orchestrator identifies a set of
relevant skills $\Sigma_t \subseteq \Sigma_{\psi_t}$ that are active for the current state.
Agent selection is then performed by aggregating competence estimates over this skill set and trading them off against execution cost:
\begin{equation*}
A_t^*
=
\operatorname*{argmax}_{A \in \mathcal{A}_{\psi_t}}
\left[
\mathbb{E}_{\sigma \in \Sigma_t}[\phi_{A,\sigma}]
-
\lambda_c \cdot \hat{C}_A(\psi_t)
\right].
\end{equation*}

where $\phi_{A,\sigma}$ is the performance estimate stored in the agent profile $\mathcal{P}_{A,\psi_t}$.
In practice, we approximate the expected competence by aggregating the posterior
means over the active skill set and optionally incorporating semantic alignment between the current state and the agent profile:
\begin{equation*}
% $
A_t^*=\operatorname*{argmax}_{A\in\mathcal{A}_{\psi_t}}
\Bigg[\underbrace{\sum_{\sigma\in\Sigma_t} w_{t,\sigma}\frac{\alpha_{A,\sigma}}{\alpha_{A,\sigma}+\beta_{A,\sigma}}}_{\text{Estimated Competence}}
-\lambda_c\cdot\underbrace{\hat{C}_A(\psi_t)}_{\text{Mode-Specific Cost}}\Bigg].
% $
\end{equation*}

This ensures that each decision is grounded in task-relevant skills, agent-specific competence estimates, and explicit cost constraints. The full algorithm is provided in Appendix~\ref{app:algm} (Algorithm~\ref{alg:routing}).

\subsection{Skill Handbook Learning}\label{sec:learning}
We construct and refine the Skill Handbook $\mathcal{H}$ from execution traces rather than learning a monolithic routing policy. The procedure iteratively updates the skill registry, agent profiles, and mode-level routing metadata (Figure~\ref{fig:skillorchestra_overview}, left).

\textbf{Phase 1: Skill Discovery and Profile Construction.}
We assume an exploratory dataset
$\mathcal{D}_{\text{train}} = \{ (q_i, \mathcal{B}_i) \}_{i=1}^{N}$,
where $\mathcal{B}_i=\{\tau_i^{(1)},\tau_i^{(2)},\ldots\}$ are trajectories obtained by varying the agent choice at specific modes.

For each query and mode $\psi$, we contrast a successful trajectory $\tau^{\psi}_{+}$ with a failed one $\tau^{\psi}_{-}$. Their difference $\mathcal{D}_{\text{diff}}(\tau^{\psi}_{+} \parallel \tau^{\psi}_{-})$ isolates the missing capability.
An LLM-based discoverer abstracts this capability gap into a reusable skill definition $\sigma_{\text{new}}$, which is added to the registry $\mathcal{V}_{\Sigma}$ together with its associated mode mapping ${M}$.

Agent profiles are then estimated from aggregated outcomes. For each agent $A$, mode $\psi$, and skill $\sigma \in \Sigma_{\psi}$, we model success probability as
$\phi_{A,\sigma} \sim \mathrm{Beta}(\alpha_{A,\sigma}, \beta_{A,\sigma})$,
updated via
\begin{equation*}
\begin{aligned}
\alpha_{A,\sigma}^{(t+1)} &\leftarrow \alpha_{A,\sigma}^{(t)} +
\sum_{\tau \in \mathcal{B}_i} \mathbb{I}[A \text{ succeeds on } \sigma \text{ in } \tau], \\
\beta_{A,\sigma}^{(t+1)} &\leftarrow \beta_{A,\sigma}^{(t)} +
\sum_{\tau \in \mathcal{B}_i} \mathbb{I}[A \text{ fails on } \sigma \text{ in } \tau].
\end{aligned}
\end{equation*}
Mode-level routing signals (e.g., frequent transitions or systematic failures or recurring recovery patterns) are distilled into reusable mode-selection insights and stored as routing metadata $\mathcal{R}_{\psi}$.

\textbf{Phase 2: Handbook Refinement.}
To prevent over-fragmentation or redundancy, we periodically refine the skill set using agent profile statistics.

\ding{70} \textbf{Splitting.}
A skill $\sigma$ is marked as a split candidate if agent performance exhibits high variance across its associated queries, indicating multiple underlying capabilities.

\ding{70} \textbf{Merging.}
A pair of skills $(\sigma_i,\sigma_j)$ is marked as a merge candidate when their agent performance profiles are statistically indistinguishable, suggesting redundancy for routing.

Given these candidates, an LLM-based reflector (e.g., GPT-5) reviews the proposed operations and, if appropriate, generates revised skill definitions. Approved refinements update both the skill registry and the associated competence statistics $(\alpha_{A,\sigma}, \beta_{A,\sigma})$.
The final refined handbook $\mathcal{H}^*$ encodes learned skills, agent profiles, and routing metadata, and serves as the reusable knowledge base for inference-time handbook selection (Section~\ref{sec:inference}).

\subsection{Pareto-Optimal Skill Handbook Selection}
\label{sec:selection}
This subsection formalizes the handbook selection step introduced in
Section~\ref{sec:inference}, where an orchestrator-specific subset is chosen
to match the reasoning capacity and cost budget of the target orchestrator. An illustration can be found in Figure~\ref{fig:skillorchestra_overview} (middle).

Given the learned handbook $\mathcal{H}^*$ (Section~\ref{sec:learning}),
our goal is to select, for a target orchestrator $\mathcal{O}$,
a subset $\mathcal{H} \subseteq \mathcal{H}^*$ that achieves the best
end-to-end performance-cost tradeoff.

Each candidate subset $\mathcal{H}$ induces a routing policy
$\pi_{\mathcal{H}}$, which produces a trajectory
$\tau_{\mathcal{H}}(q)$ for a query $q$.
We evaluate candidate subsets on a held-out validation set
$\mathcal{D}_{\text{val}}$ and solve:
\begin{equation*}
\mathcal{H}_{\text{base}}^{(\mathcal{O})}
=
\operatorname*{argmax}_{\mathcal{H} \subseteq \mathcal{H}^*}
\mathbb{E}_{q \sim \mathcal{D}_{\text{val}}}
\left[
R(\tau_{\mathcal{H}}(q))
-
\lambda \sum_{t=0}^{|\tau_{\mathcal{H}}(q)|}
C(\psi_t, A_t)
\right].
\end{equation*}
Here, $R(\tau_{\mathcal{H}}(q)) \in [0,1]$ denotes task success,
and $C(\psi_t, A_t)$ is the execution cost at step $t$.
The coefficient $\lambda$ controls the performance-cost tradeoff.
This objective directly evaluates entire trajectories rather than
local routing accuracy, ensuring that the selected handbook lies on
the Pareto frontier for the target orchestrator.

\begin{table*}[t!]
\footnotesize
\centering
\caption{Experimental results on QA datasets. \best{Bold} = best, \second{underline} = second best in each column. SkillOrchestra uses the same orchestrator model as baselines.
SkillOrchestra+ reports the best performance obtained by switching among different orchestrator models within the same agent pool while using the same learned Skill Handbook.}
\vspace{-0.6em}
\label{tab:qa_em_routing_groups}
\resizebox{0.90\linewidth}{!}{
\begin{tabular}{l ccc cccc c}
\toprule
\textbf{Method} &
\multicolumn{3}{c}{\textbf{General QA}} &
\multicolumn{4}{c}{\textbf{Multi-Hop QA}} &
\textbf{Avg.} \\
\cmidrule(lr){2-4}\cmidrule(lr){5-8}
& \textbf{NQ} & \textbf{TriviaQA} & \textbf{PopQA} & \textbf{HotpotQA} & \textbf{2wiki} & \textbf{Musique} & \textbf{Bamboogle} & \\
\midrule
Vanilla    & 9.2  & 26.0 & 12.2 & 14.0 & 26.6 & 2.6  & 4.0  & 13.5 \\
\midrule
\rowcolor{myred!20}
\multicolumn{9}{c}{\textbf{\textit{No Routing}}} \\
SFT       & 21.2 & 40.0 & 16.0 & 19.8 & 25.6 & 5.2  & 11.2 & 19.9 \\
RAG       & 29.8 & 54.0 & 36.6 & 21.6 & 14.6 & 7.8  & 22.4 & 26.7 \\
CoT~\citep{wei2022chain}       & 12.6 & 35.8 & 16.0 & 16.8 & 20.8 & 4.6  & 22.4 & 18.4 \\
Search-R1~\citep{jin2025searchr} & 32.8 & 51.0 & 32.4 & 23.6 & 27.8 & 9.0  & 27.2 & 29.1 \\
\midrule

\rowcolor{myyellow!20}
\multicolumn{9}{c}{\textbf{\textit{Heuristic \& Discriminative Routing}}} \\
Largest LLM  & 29.6 & 57.8 & 35.4 & 27.8 & 27.4 & 10.4 & 48.0 & 33.8 \\
Prompt LLM   & 30.0 & 58.0 & 34.0 & 26.8 & 26.2 & 10.8 & 44.8 & 32.9 \\
Prompt LLM+ (multi turn)  & 25.8 & 50.0 & 25.6 & 20.6 & 24.8 & 7.8  & 47.2 & 28.8 \\
KNN Router~\citep{hu2024routerbench}   & 26.2 & 52.8 & 22.2 & 22.4 & 19.6 & 6.6  & 36.0 & 26.5 \\
KNN Router+ (multi turn)  & 23.6 & 47.8 & 23.2 & 15.4 & 23.4 & 7.2  & 38.4 & 25.6 \\
MLP Router~\citep{hu2024routerbench}   & 25.2 & 46.0 & 22.2 & 19.8 & 21.0 & 7.2  & 36.0 & 25.3 \\
BERT Router~\citep{ong2025routellm}  & 23.0 & 51.6 & 19.2 & 21.6 & 20.6 & 5.8  & 31.2 & 24.7 \\
RouterDC~\citep{chen2024routerdc}     & 27.8 & 59.2 & 28.2 & 24.4 & 21.8 & 8.0  & 50.4 & 31.4 \\
GraphRouter~\citep{feng2025graphrouter}  & 27.6 & 58.6 & 28.0 & 23.4 & 18.0 & 7.6  & 44.8 & 29.7 \\
FrugalGPT~\citep{chen2024frugalgpt}    & 26.5 & 56.2 & 36.2 & 23.4 & 26.8 & 10.3 & 43.0 & 31.8 \\
\midrule

\rowcolor{myyellow!20}
\multicolumn{9}{c}{\textbf{\textit{RL-based Routing}}} \\
Router-R1~\citep{zhang2025routerr} & 38.8 & 70.6 & 38.4 & 35.2 & 43.4 & 13.8 & 51.2 & 41.6 \\
\midrule

\rowcolor{myblue!20}
\multicolumn{9}{l}{\textbf{Ours}} \\
\ourmethod &
\second{54.2} & \second{71.6} & \second{42.6} & \second{39.0} & \second{48.0} & \second{18.2} & \second{58.4} & \second{47.4} \\
\ourmethod+ &
\best{54.8} & \best{80.2} & \best{48.8} & \best{44.2} & \best{49.6} & \best{20.6} & \best{63.2} & \best{51.6} \\
\bottomrule
\end{tabular}
}
\vspace{-1em}
\end{table*}

\section{Experiments}\label{sec:exp}
We conduct extensive experiments to answer:

\textbf{(RQ1) Effectiveness:}  
Does a learned Skill Handbook improve end-to-end accuracy over heuristic, discriminative, and RL-based methods?

\textbf{(RQ2) Efficiency:}  
Does skill-based orchestration yield a better performance-cost trade-off?

\textbf{(RQ3) Routing Behavior:}  
Does skill-based orchestration reduce routing collapse and better match model capacity to task difficulty across modes?

\textbf{(RQ4) Transferability:}  
Can a Skill Handbook be reused across orchestrators without retraining?

\textbf{(RQ5) Component Contribution:} How do different components of the Skill Handbook contribute to overall performance and cost efficiency?

\subsection{SkillOrchestra for Model Routing}

We first evaluate SkillOrchestra in the model routing setting~\citep{chen2024frugalgpt,feng2025graphrouter,zhang2025routerr}, where no external tools or knowledge base are provided. Therefore, the performance gaps directly reflect the quality of model orchestration.

\textbf{Benchmarks.} We consider a diverse suite of knowledge and reasoning-intensive tasks including (1) General QA: Natural Question~\citep{kwiatkowski2019natural}, TriviaQA~\citep{joshi-etal-2017-triviaqa}, PopQA~\citep{mallen-etal-2023-trust}; (2) Multi-hop QA: HotpotQA~\citep{yang2018hotpotqa}, 2WikiMultiHopQA~\citep{ho-etal-2020-constructing}, Musique~\citep{trivedi-etal-2022-musique}, and Bamboogle~\citep{press-etal-2023-measuring}; (3) Math Reasoning: MATH~\citep{hendrycksmath2021} and AMC23~\citep{maa_amc_2023}.

\textbf{Experimental setup and baselines.} We use Qwen2.5-3B~\citep{qwen2.5} as the orchestrator and adopt the same configuration as Router-R1 for controlled comparison with all routing baselines. Model pool and implementation details are included in Appendix~\ref{app:exp_model_route}. We compare SkillOrchestra against three categories of methods: \textbf{(1) No routing}: methods that do not dynamically consult different models, including supervised finetuning, RAG as in~\citet{zhang2025routerr}, CoT~\citep{wei2022chain}, and Search-R1~\citep{jin2025searchr};
\textbf{(2) Heuristic \& Discriminative routing}: methods that select models based on input-level signals or learned classifiers, including Largest LLM, Prompt LLM, Prompt LLM+ (explicit task decomposition+multi-turn), KNN Router~\citep{hu2024routerbench}, KNN Router+ (explicit task decomposition and route each subtask to different models which matching query similarity with KNN router), MLP Router~\citep{hu2024routerbench}, BERT Router~\citep{ong2025routellm}, RouterDC~\citep{chen2024routerdc}, GraphRouter~\citep{feng2025graphrouter}, and FrugalGPT~\citep{chen2024frugalgpt};
\textbf{(3) RL-based routing}: Router-R1~\citep{zhang2025routerr}, a strong PPO-trained~\citep{schulman2017proximal} multi-turn router with 14k samples, which represents the current SoTA in learned end-to-end orchestration. 

\begin{wrapfigure}{r}{0.5\textwidth} % {r} for right, {l} for left
    \centering
    \vspace{-8pt} % Pulls the figure up to align with the text start
    \includegraphics[width=0.5\textwidth]{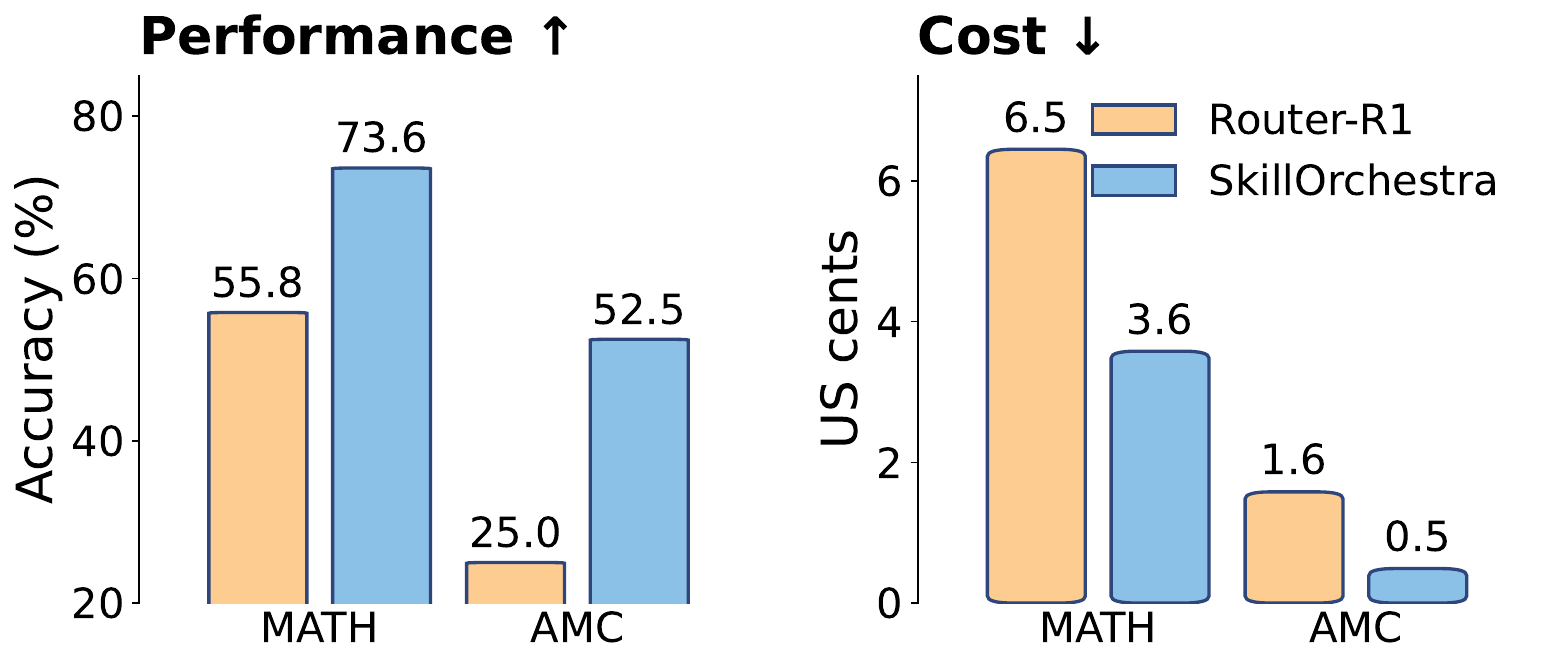}
    \vspace{-15pt}
    \caption{Performance and cost comparison: SkillOrchestra vs. Router-R1. SkillOrchestra achieves up to a 22.5 percentage-point improvement in accuracy while reducing inference cost by $\sim2.0\times$.}
    \label{fig:math}
    \vspace{-9pt} % Adjust bottom spacing to pull text closer
\end{wrapfigure}

\textbf{Observation \ding{182} SkillOrchestra outperforms all routing baselines, including expensive RL-based methods (RQ1).} 
SkillOrchestra surpasses all baselines on both general and multi-hop QA (Table~\ref{tab:qa_em_routing_groups}). Compared to Router-R1 (41.6 EM), SkillOrchestra reaches 47.4 (+5.8), and SkillOrchestra+ achieves 51.6 (+10.0). Gains are especially large on multi-hop tasks such as Musique (13.8 → 18.2 → 20.6) and Bamboogle (51.2 → 58.4 → 63.2). Similar trends hold for math reasoning (Figure~\ref{fig:math}), with up to +22.5 accuracy over Router-R1 at substantially lower cost. Notably, these gains require only a small fraction of the training data, demonstrating higher data efficiency than RL-based routing.

\textbf{Observation \ding{183} SkillOrchestra lies on the Pareto frontier (RQ2).} 
Figure~\ref{fig:pareto_overview} (left) shows that SkillOrchestra and SkillOrchestra+ achieve higher accuracy at lower or comparable cost than all heuristic, discriminative, and RL-based baselines.
Importantly, higher per-token price does not necessarily imply higher total inference cost. 
Total cost depends jointly on (i) the per-token price of the selected backbone model, (ii) the number of generated tokens, and (iii) the number of routing steps. 
In practice, some lower per-token models produce substantially longer reasoning chains, leading to higher overall cost.
SkillOrchestra explicitly accounts for this trade-off, often selecting capable yet more cost-efficient models (e.g., Mixtral-8×22B) instead of consistently escalating to the most expensive model (LLaMA-3.1-70B). 
For example, Router-R1 attains 41.6 EM at a high cost (51.8¢), whereas SkillOrchestra achieves higher accuracy (47.4 EM) at a lower cost (38.4¢). SkillOrchestra+ further improves to 51.6 EM at 41.6¢.
Router-R1 attains 41.6 EM at a high cost (51.8¢), whereas SkillOrchestra achieves higher accuracy (47.4 EM) at a lower cost (38.4¢). SkillOrchestra+ further improves to 51.6 EM at 41.6¢.
Similar advantages appear in math reasoning (Figure~\ref{fig:math}), where SkillOrchestra improves accuracy while reducing cost by about $2\times$. These results indicate that skill-aware routing allocates models more efficiently and shortens reasoning chains.

\begin{figure}[t!]
    \centering
    \includegraphics[width=1\linewidth]{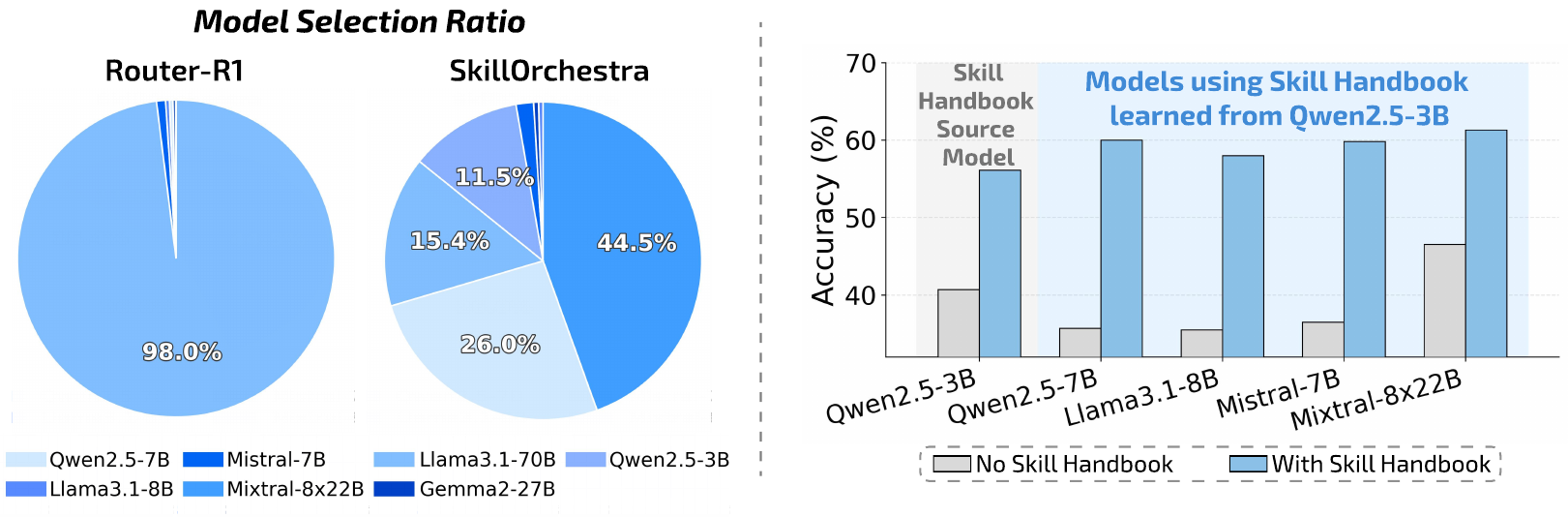}
    \caption{Skill-based orchestration mitigates routing collapse and generalizes across orchestrators. (Left) Router-R1 collapses to a single large model (98\% Llama3.1-70B), while SkillOrchestra distributes calls according to capability differences. (right) A Skill Handbook learned from Qwen2.5-3B transfers across orchestrator backbones without retraining, consistently improving performance and achieving larger gains with stronger backbones.}
    \label{fig:model_select_transfer}
\end{figure}

\textbf{Observation \ding{184} Skill-based routing alleviates routing collapse seen in RL-based routing (RQ3).}
To understand the performance and efficiency gap, Figure~\ref{fig:model_select_transfer} (left) compares model selection distributions across nine benchmarks. Router-R1 shows clear \textit{routing collapse}: it selects LLaMA-3.1-70B for 98.02\% of all calls, while all other models are almost unused (each $\leq$0.92\%; e.g., Qwen2.5-7B 0.35\%, Mistral-7B 0.92\%, Mixtral-8×22B 0.04\%, Qwen2.5-3B 0.00\%). Despite being trained as a multi-model router, its RL policy converges to repeatedly invoking a single large model, limiting specialization and inflating cost.
In contrast, \ourmethod produces a much more balanced routing pattern: e.g., Mixtral-8×22B 44.53\%, Qwen2.5-7B 25.99\%, LLaMA-3.1-70B 15.38\%, and Qwen2.5-3B 11.50\%. This distribution reflects capability-aware specialization, where stronger models are used only when necessary and lighter models handle simpler steps.
Importantly, skill-based routing also makes the orchestrator itself more effective. In some cases, the orchestrator can directly answer the query without escalating to a larger model, further reducing unnecessary calls and lowering the total cost. An example in shown in Figure~\ref{fig:trace_no_help_amc}.

\textbf{Observation \ding{185} The learned skill handbook transfers across orchestrator backbones without retraining (RQ4).} We reuse the skill handbook learned from traces where Qwen2.5-3B serves as the orchestrator, and directly apply it to other backbone models without any additional handbook training. Figure~\ref{fig:model_select_transfer} (right) shows performance before and after introducing the same skill handbook, with results averaged over three general QA datasets.
The learned handbook consistently improves all tested models. Qwen2.5-3B itself improves from 40.7\% to 56.1\% (+15.4). When transferred to larger or stronger models, the gains remain substantial: Qwen2.5-7B improves from 35.7\% to 60.0\% (+24.3), Llama3.1-8B from 35.5\% to 58.0\% (+22.5), and Mistral-7B from 36.5\% to 59.8\% (+23.3). Even larger-scale models benefit from such handbook: Mixtral-8x22B from 46.5\% to 61.3\% (+14.8).
These results show that the skill handbook captures transferable, model-agnostic orchestration knowledge. Notably, stronger models often achieve the highest absolute performance when paired with the transferred handbook, suggesting that improved backbone capability and structured skill guidance are complementary.

\subsection{SkillOrchestra on Agent Orchestration} 
We next evaluate whether SkillOrchestra extends beyond model routing to full agent orchestration, where the system must coordinate multiple operational modes and tools beyond model selection. We use the same configuration as ToolOrchestra~\citep{toolorchestra}, detailed in the following. 

\textbf{Experimental setup and baselines.}
We evaluate on FRAMES~\citep{krishna2024factfetchreasonunified} and consider three operational modes: \texttt{search} (web and local search), \texttt{code}, and \texttt{answer}. Each mode corresponds to a different model pool, detailed in Appendix~\ref{app:exp_agent_orch}. The maximum interaction horizon is 50 turns. With Qwen3-8B as the orchestrator, we compare against ToolOrchestra~\citep{toolorchestra}, which trains the orchestrator using GRPO. We also compare against strong proprietary model orchestrators such as GPT-5~\citep{gpt5}, Gemini-3-Pro~\citep{google_gemini_3}, or Claude-Opus-4.5~\citep{anthropic_claude_opus_4_5}, while keeping modes, model pools, tools, and execution environments fixed.

\begin{wraptable}{r}{0.5\textwidth} % {r} for right side, {l} for left
\footnotesize
\centering
\caption{Analysis of SkillOrchestra’s Skill Handbook design. HB: Has Handbook, Disc: Skill Discovery, Ref: Handbook Refinement, Sel: Handbook Selection, FG: Fine-grained Skills. Orchestrator: Qwen3-8B.}
\setlength{\tabcolsep}{3.5pt}
\renewcommand{\arraystretch}{1.1}
\begin{tabular}{lccccc@{\hspace{5pt}}cc}
\rowcolor{myblue!40}
\textbf{Setting} & \textbf{HB} & \textbf{Disc} & \textbf{Ref} & \textbf{Sel} & \textbf{FG} & \textbf{Acc \%} & \textbf{Cost \$} \\
\Xhline{1pt}
No HB & \xmark & \xmark & \xmark & \xmark & \xmark & 71.0 & 122.9 \\
No Ref + Sel & \cmark & \cmark & \xmark & \xmark & \cmark & 79.0 & 5.5 \\
No Selection & \cmark & \cmark & \cmark & \xmark & \cmark & 79.3 & 3.4 \\
No FG Skills & \cmark & \cmark & \cmark & \cmark & \xmark & 80.4 & 15.1 \\
\midrule
Full System & \cmark & \cmark & \cmark & \cmark & \cmark & 85.0 & 9.3 \\
\bottomrule
\end{tabular}
\label{tab:ablation}
\vspace{-10pt}
\end{wraptable}

\textbf{Observation \ding{186} SkillOrchestra achieves better performance-cost trade-offs in full agent orchestration (RQ1, RQ2).}
Figure~\ref{fig:pareto_overview} (right) shows that SkillOrchestra remains on the Pareto frontier in the more complex agent orchestration setting with multiple tools, operational modes beyond model routing. Our method achieves the highest accuracy (84.3\%) while also incurring the lowest total cost (\$72.7) among strong learned and proprietary-model baselines.
Compared to the RL-trained ToolOrchestra (76.3\%, \$92.7), SkillOrchestra improves accuracy by +8.0 points while reducing cost by 21.6\%. It also outperforms stronger proprietary orchestrators such as GPT-5 (74.6\%, \$120.4), Claude Opus 4.5 (77.9\%, \$758.1), and Gemini 3 Pro (78.9\%, \$1729.3).
These results highlight an important system-level trade-off: while using a stronger model as the orchestrator can improve raw task performance, it often does so at a prohibitive cost due to expensive per-token pricing and long multi-step trajectories. In contrast, SkillOrchestra improves both accuracy and efficiency by coordinating specialized models and tools through explicit skill modeling, rather than relying on a single large model to carry the entire process.

\textbf{Observation \ding{187} More skills are not always better; optimal performance-cost trade-offs require refining and selecting skills to match the orchestrator’s capability (RQ1, RQ3, RQ5).} 
To understand the contribution of each component of SkillOrchestra, we conduct a controlled ablation study on 100 randomly sampled FRAMES tasks. 
As shown in Table~\ref{tab:ablation}, removing the Skill Handbook causes a large drop in accuracy (85.0\% → 71.0\%) and a sharp increase in cost (9.3 → 122.9), showing that structured skill guidance is crucial for both effectiveness and efficiency.
Using discovered skills without handbook refinement and selection, which includes redundant, overlapping or overly broad skills, still achieves reasonable accuracy (79.0\%) at low cost (5.5), suggesting that even an unrefined skill set still provides useful routing signals. 
Enabling refinement further reduces cost (3.4) while maintaining similar accuracy (79.3\%), indicating that reorganizing skills by merging redundant ones and splitting indistinguishable ones improves efficiency.
Disabling fine-grained skills degrades both accuracy (80.4\%) and efficiency (15.1), showing that appropriately detailed skills help the orchestrator make better decisions.
Overall, the best performance-cost trade-off is achieved when skills are discovered, reorganized, and selectively applied at a level of detail that the orchestrator can use effectively.

\section{Conclusion}\label{sec:conclude}
In this work, we propose SkillOrchestra, an agentic orchestration framework that reframes multi-turn routing as skill-grounded decision making. By learning a Skill Handbook, the orchestrator makes state-aware, competence-aware decisions that explicitly optimize the performance-cost trade-off. Across both model routing and agent orchestration settings, SkillOrchestra achieves superior performance with significantly lower cost compared to competitive baselines. Moreover, the handbook is transferable across orchestrator backbones without retraining, enabling scalable deployment as model pools evolve. We hope this work serves as a springboard for scalable orchestration that improves the performance-cost frontier as agent pools grow and diversify.

\bibliography{main}

%%%%%%%%%%%%%%%%%%%%%%%%%%%%%%%%%%%%%%%%%%%%%%%%%%%%%%%%%%%%%%%%%%%%%%%%%%%%%%%
%%%%%%%%%%%%%%%%%%%%%%%%%%%%%%%%%%%%%%%%%%%%%%%%%%%%%%%%%%%%%%%%%%%%%%%%%%%%%%%
% APPENDIX
%%%%%%%%%%%%%%%%%%%%%%%%%%%%%%%%%%%%%%%%%%%%%%%%%%%%%%%%%%%%%%%%%%%%%%%%%%%%%%%
%%%%%%%%%%%%%%%%%%%%%%%%%%%%%%%%%%%%%%%%%%%%%%%%%%%%%%%%%%%%%%%%%%%%%%%%%%%%%%%
\newpage
\appendix
\onecolumn
\section{Experimental Details}\label{app:exp}

\subsection{Experimental Details for Model Routing}\label{app:exp_model_route}
\textbf{Implementation Details.} 
We use the same evaluation protocol as Router-R1 for controlled comparison with all routing baselines.
We use Qwen2.5-3B~\citep{qwen2.5} as the orchestrator, and the model pool consists of Qwen2.5-7B~\citep{qwen2.5}, LLaMA-3.1-8B~\citep{grattafiori2024llama3herdmodels}, LLaMA-3.1-70B~\citep{grattafiori2024llama3herdmodels}, Mistral-7B~\citep{jiang2023mistral7b}, Mixtral-8x22B~\citep{jiang2024mixtral}, and Gemma-2-27B~\citep{team2024gemma}. Routing operates in two modes: (1) search mode, where the orchestrator selects a model from the pool to perform subtasks (provide knowledge or solve the subtask); and (2) answer mode, where the orchestrator aggregates intermediate results and produce the final answer. We set the max number of turns to 4. We evaluate the performance using Exact Match (EM) and efficiency using total completion cost. SkillOrchestra is trained in a low-data regime: by default, we select $k$ ($k<50$) samples from each dataset to train the Skill Handbook and $k$ additional samples for validation and handbook retrieval. We use SkillOrchestra+ to denote the best performance obtained by switching among different orchestrator models within the same agent pool while using the same learned Skill Handbook.
\subsection{Experimental Details for Agent Orchestration}\label{app:exp_agent_orch}
\textbf{Implementation Details.} 
We follow the same evaluation protocol and experimental setup as ToolOrchestra to ensure a controlled and comparable evaluation.
We consider three operational modes:
For $\psi=\texttt{search}$, the allowable tools are
${T}_{\texttt{search}} = \{\texttt{WebSearch},\ \texttt{LocalSearch}\}$,
where WebSearch uses the Tavily API and LocalSearch uses a FAISS index built with Qwen3-Embedding-8B~\citep{qwen3embedding}. The model set is
$\mathcal{M}_{\texttt{search}} = \{\text{GPT-5},\ \text{GPT-5-mini},\ \text{Qwen3-32B}\}$.
Valid agents are compositions $(m, {T}_{\texttt{search}})$ with $m \in \mathcal{M}_{\texttt{search}}$.
For $\psi=\texttt{code}$, the tool set is
${T}_{\texttt{code}} = \{\texttt{PythonExec}\}$ operating in a sandbox, and
$\mathcal{M}_{\texttt{code}}$ = \{\text{GPT-5},\ \text{GPT-5-mini},\ \text{Qwen2.5-Coder-32B}\}.
Valid agents are $(m, T_{\texttt{code}})$ with $m \in \mathcal{M}_{\texttt{code}}$.
For $\psi=\texttt{answer}$, no external tools are used
($T_{\texttt{answer}}=\varnothing$), and
$\mathcal{M}_{\texttt{answer}}$ = \{\text{GPT-5},\ \text{GPT-5-mini},\ \text{Llama-3.3-70B-Instruct},\ \text{Qwen3-32B},\ \text{Qwen2.5-Math-72B},\ \text{Qwen2.5-Math-7B}\}.
Valid agents are $(m, \varnothing)$ with $m \in \mathcal{M}_{\texttt{answer}}$. 
The maximum interaction horizon is 50 turns. Final answers are evaluated for accuracy using GPT-5-mini as a judge, and total system cost (USD) is measured.

\section{Skill-Grounded Agent Routing Algorithm Pseudocode}\label{app:algm}
We present an algorithm block for Skill-grounded Agent Routing in Algorithm~\ref{alg:routing}. A concrete illustration can be found in Figure~\ref{fig:skillorchestra_overview} (Deployment).
\begin{algorithm}[!h]
\caption{Skill-Grounded Agent Routing by Orchestrator $\mathcal{O}$}
\label{alg:routing}
\small
\DontPrintSemicolon
\Input{State $s_t$;  query handbook $\mathcal{H}_q$; cost weight $\lambda_c$}
\Output{Selected mode $\psi_t$, agent $A_t$, trace $z_t$, observation $o_t$, updated state $s_{t+1}$}

\BlankLine
\Comment{\textbf{Mode selection}}
Select operational mode $\psi_t \sim \pi_{\text{mode}}(\cdot \mid s_t;\,\mathcal{R}_{\psi})$

\BlankLine
\Comment{\textbf{Retrieve active skills}}
Retrieve active skills $\Sigma_t \subseteq \Sigma_{\psi_t}$ from $\mathcal{H}_q$\;

\BlankLine
\Comment{\textbf{Competence-aware routing}}
\ForEach{$A \in \mathcal{A}_{\psi_t}$}{
  \Comment{posterior-mean competence from estimated stats in the Handbook}
  $\widehat{P}(A) \leftarrow \sum_{\sigma \in \Sigma_t} w_{t,\sigma}\,
    \dfrac{\alpha_{A,\sigma}}{\alpha_{A,\sigma}+\beta_{A,\sigma}}$\;

  \Comment{utility = competence - mode-specific cost}
  $U(A) \leftarrow \widehat{P}(A) - \lambda_c \cdot \widehat{C}_A(\psi_t)$\;
}

$A_t \leftarrow \arg\max_{A\in\mathcal{A}_{\psi_t}} U(A)$\;

\BlankLine
\Comment{\textbf{Execute + state transition}}
$(z_t,o_t) \leftarrow \mathrm{Execute}(A_t,\psi_t,s_t)$ \tcp*{$z_t$ = agent trace, $o_t$ = env observation}
$s_{t+1} \leftarrow \mathrm{UpdateState}(s_t,\psi_t,A_t,z_t,o_t)$\;

\end{algorithm}

\section{A Closer Look at Model Selection: SkillOrchestra vs. ToolOrchestra}~\label{app:add_res}

\textbf{Skill-grounded routing leads to more efficient tool-model allocation (RQ3).}
To understand the benefits of SkillOrchestra compared to ToolOrchestra, we also take a closer look at the model selection ratio at each operational mode.
We found that the cost reduction of SkillOrchestra comes from smarter allocation of models across different operational modes, rather than simply reducing the number of calls.
In search mode, ToolOrchestra routes 99.7\% of calls to GPT-5-mini, whereas SkillOrchestra instead uses Qwen3-32B (also the cheapest) for 100\% of search calls, identifying it as sufficiently capable and more cost-efficient for the search task.
In answer mode, ToolOrchestra similarly exhibits routing collapse, routing 97.9\% of calls to GPT-5. SkillOrchestra distributes answer generation more strategically: GPT-5 is used in 58.4\% of calls, with the remainder handled by cheaper or specialized models such as GPT-5-mini (10.0\%) and Qwen3-32B or math-expert models.
This diversification allows the system to reserve expensive models for truly difficult reasoning steps while offloading simpler synthesis or domain-specific subtasks to more efficient models. 

\section{Demonstrations of Skill-Aware Orchestration}
\label{app:demo_traces}

We provide full execution traces of the skill-based router in 
Figures~\ref{fig:example_multi_amc}--\ref{fig:router_trace_popqa_noisy}, 
along with the orchestration instruction template in 
Figure~\ref{fig:agent_orch_inst}. 
The instruction integrates the task query, execution context, and the selected Skill Handbook used for routing decisions.

% --- listing config (page-break & long prompt friendly) ---
\lstset{
  basicstyle=\ttfamily\scriptsize,
  breaklines=true,
  columns=fullflexible,
  keepspaces=true,
  showstringspaces=false,
  aboveskip=0pt,
  belowskip=0pt,
  xleftmargin=0pt,
  frame=none
}

\begin{tcolorbox}[
  title={AMC Example: Skill-Based Router Corrects via Multi-turn Routing},
  sharp corners,
  breakable,
  colframe=Periwinkle,
  colback=white,
  boxrule=2.2pt,
  boxsep=0.9pt,
  enhanced,
  shadow={3pt}{-3pt}{0pt}{opacity=1,mygrey},
  % subtle transcript feel (optional)
  borderline west={1.1pt}{0pt}{Periwinkle!65},
]

% --- key properties strip ---
\noindent
\statuspill{TagGreen}{\ding{51}\ Skill Handbook}\hspace{0.55em}
\statuspill{TagOrange}{2 external model calls (\texttt{<search>})}\hspace{0.55em}
\statuspill{TagGreen}{\ding{51}\ Correct}\hspace{0.55em}
\vspace{0.45em}

{\scriptsize
\begin{lstlisting}[language={}, escapeinside={(*@}{@*)}]

(*@\tagpill{cyan}{Skill Router Instruction}@*)
You are a skill-based model router. You are selecting the best model to answer a question by analyzing a question to identify required skills and their importance related to this question.

## Learned Skill Definitions (from validation)

### Algebra and Functions
Symbolic manipulation and equation-solving across rational expressions, logarithms/exponents, polynomials via Vieta, and trigonometric identities and parameters.
  - Rational/linear manipulation (nested fractions, clearing denominators, simple systems): Evaluate nested or continued fractions, reduce to irreducible form, and set up/solve linear relations from worded constraints.
    Examples: Problem 1: Compute 3 + 1/(3 + 1/(3 + 1/3)) as an irreducible fraction and return m+n., Problem 12: Three numbers sum to 96 with linear relations; find |first - second|.
  - Logarithm/exponent identities and metric constraints: Use log laws to simplify expressions and translate distance conditions on the number line into equations in log variables.
    Examples: Problem 3: Distance between log_6 x and log_6 9 equals twice another distance; find product of solutions., Problem 6: Evaluate (log 5)^3 + (log 20)^3 + (log 8)(log 0.25).
  - Vieta and symmetric sums with parameter shifts: Extract sums/products of roots from polynomial coefficients and evaluate expressions after shifting variables (e.g., edges increased by a constant).
    Examples: Problem 7: Roots are box dimensions; edges lengthened by 2; compute new volume via Vieta.
  - Trigonometric identity reduction and parameter interval analysis: Rewrite trig sums/products (e.g., sin x + sin 2x) and determine parameter ranges guaranteeing multiple solutions or specific solution behaviors.
    Examples: Problem 9: For a*(sin x + sin 2x) = sin 3x, find all a producing more than one solution and compute p+q+r.

### Geometry and Transformations
Spatial reasoning in 2D/3D using vectors, coordinates, isometries, and complex-plane interpretations of loci and areas.
  - 3D vector/coordinate geometry and dot products: Model regular solids with coordinates, use midpoints and vectors, and compute angles via dot products and norms.
    Examples: Problem 4: In regular tetrahedron ABCD with M midpoint of AB, find $\cos(\angle CMD)$ as p/q.
  - Complex-plane geometry: loci, polygons, and area optimization: Interpret complex constraints as geometric loci (segments, disks), compute Minkowski sums (stadium/rounded rectangle areas), and analyze polygons formed by z and 1/z under quadratic relations.
    Examples: Problem 5: Region from sum of a segment [3, 4i] and a unit disk; find closest integer to area., Problem 14: For z^2 - cz + 10 = 0, quadrilateral with z1, z2, 1/z1, 1/z2 has maximal area; find $c=\sqrt{m}$.
  - Coordinate geometry with distance constraints in special quadrilaterals: Place figures in coordinates, encode equal-leg/parallel-side constraints, and solve using distance equations to find side ratios.
    Examples: Problem 13: Isosceles trapezoid with PA=1, PB=2, PC=3, PD=4; find BC/AD.
  - Composition of plane isometries and periodicity: Represent rotations and reflections (matrices or angle-line representations), compose varying-parameter isometries, and determine the least n returning a point to itself.
    Examples: Problem 10: Find minimal n so $T_1\circ T_2\circ \cdots \circ T_n$ sends (1,0) back to itself.

### Number Theory and Diophantine Analysis
Integer-structure problems involving Pell equations, valuations, and divisibility properties of rational sums.
  - Pell-type equations for figurate-number squares: Convert conditions like triangular numbers being squares into Pell equations, use fundamental solutions/recurrences to generate the next solutions.
    Examples: Problem 8: Find the fourth triangular number that is also a square; sum its digits.
  - Harmonic denominators vs LCM via p-adic valuations: Analyze reduced denominators of harmonic numbers, compare to LCM(1..n), and count n where strict inequality holds using prime-power valuations and cancellation.
    Examples: Problem 15: For 1 <= n <= 22, count n with k_n < L_n.

### Combinatorics and Discrete Structures
Counting and structural reasoning for pairings and permutations under process constraints.
  - Constrained pairing/matching counts: Model pairings with inequality or dominance constraints, assess feasibility (often via greedy/ordering), and count valid matchings.
    Examples: Problem 2: Number of ways to pair 1..14 so larger >= 2x smaller in each pair.
  - Permutation process modeling (passes/runs) and counting: Translate left-to-right multi-pass selection procedures into properties like increasing runs or pile counts and enumerate permutations achieving a given number of passes.
    Examples: Problem 11: Count orderings of 13 cards that are picked up in exactly two passes.

## Model Performance (learned from validation)

### LLaMA-3.1-70B-Instruct
Overall: 15% success (3/20)
Skill scores:
  - Pell-type equations for figurate-number squares: 100%
  - Harmonic denominators vs LCM via p-adic valuations: 50%
  - Complex-plane geometry: loci, polygons, and area optimization: 33%
  - Rational/linear manipulation (nested fractions, clearing denominators, simple systems): 0%
  - Constrained pairing/matching counts: 0%
Strengths: geometry.complex_plane_loci_area
Weaknesses: geometry.3d_dot_product, algebra.rational_linear_manipulation, math-heavy reasoning in this sample

### Gemma-2-27B-Instruct
Overall: 10% success (2/20)
Skill scores:
  - Harmonic denominators vs LCM via p-adic valuations: 50%
  - Rational/linear manipulation (nested fractions, clearing denominators, simple systems): 25%
  - Constrained pairing/matching counts: 0%
  - Logarithm/exponent identities and metric constraints: 0%
  - 3D vector/coordinate geometry and dot products: 0%
Strengths: algebra.rational_linear_manipulation, basic_algebra_word_problems
Weaknesses: algebra.logs_and_exponents, quantitative reasoning in this sample, geometry.complex_plane_loci_area

### Qwen2.5-7B-Instruct
Overall: 25% success (5/20)
Skill scores:
  - 3D vector/coordinate geometry and dot products: 100%
  - Harmonic denominators vs LCM via p-adic valuations: 100%
  - Complex-plane geometry: loci, polygons, and area optimization: 33%
  - Rational/linear manipulation (nested fractions, clearing denominators, simple systems): 25%
  - Constrained pairing/matching counts: 0%
Strengths: algebra.rational_linear_manipulation, geometry.3d_dot_product (observed), general_backup
Weaknesses: algebra.logs_and_exponents, combinatorics.constrained_pairing_matching, algebra.rational_linear_manipulation

### Mistral-7B-Instruct
Overall: 25% success (5/20)
Skill scores:
  - Harmonic denominators vs LCM via p-adic valuations: 100%
  - Logarithm/exponent identities and metric constraints: 50%
  - Coordinate geometry with distance constraints in special quadrilaterals: 50%
  - Rational/linear manipulation (nested fractions, clearing denominators, simple systems): 25%
  - Constrained pairing/matching counts: 0%
Strengths: algebra.logs_and_exponents (partial), geometry.coordinate_distance_quadrilaterals, algebra.rational_linear_manipulation
Weaknesses: geometry.3d_dot_product, algebra.rational_linear_manipulation, combinatorics.permutation_passes_runs

### Mixtral-8x22B-Instruct
Overall: 20% success (4/20)
Skill scores:
  - Pell-type equations for figurate-number squares: 100%
  - Rational/linear manipulation (nested fractions, clearing denominators, simple systems): 50%
  - Logarithm/exponent identities and metric constraints: 50%
  - Constrained pairing/matching counts: 0%
  - 3D vector/coordinate geometry and dot products: 0%
Strengths: algebra.logs_and_exponents, algebra.rational_linear_manipulation, general algebraic manipulation
Weaknesses: combinatorics.constrained_pairing_matching (no clear advantage), general, geometry.coordinate_distance_quadrilaterals

### LLaMA-3.1-8B-Instruct
Overall: 20% success (4/20)
Skill scores:
  - Vieta and symmetric sums with parameter shifts: 100%
  - Logarithm/exponent identities and metric constraints: 50%
  - Harmonic denominators vs LCM via p-adic valuations: 50%
  - Rational/linear manipulation (nested fractions, clearing denominators, simple systems): 25%
  - Constrained pairing/matching counts: 0%
Strengths: algebra.logs_and_exponents (partial), algebra.rational_linear_manipulation
Weaknesses: algebra.rational_linear_manipulation, geometry.coordinate_distance_quadrilaterals, combinatorics.permutation_passes_runs

## Cost Tiers (cheapest to most expensive)
- Cheap: Qwen2.5-7B-Instruct, LLaMA-3.1-8B-Instruct, Mistral-7B-Instruct
- Medium: Gemma-2-27B-Instruct
- Expensive: LLaMA-3.1-70B-Instruct, Mixtral-8x22B-Instruct

## Task
1. First, analyze the question below and identify which skills are needed, along with the percentage/weight of each skill (how important each skill is for answering this question).

Output your analysis in the following format inside <skill_analysis> tags:
<skill_analysis>
{
  "required_skills": [
    {"skill_id": "skill.id", "percentage": 50},
    {"skill_id": "skill.id", "percentage": 30},
    ...
  ],
  "reasoning": "Brief explanation of why these skills are needed"
}
</skill_analysis>

The percentages should sum to approximately 100 (they don't need to be exact, but should reflect relative importance).

2. After providing the skill analysis, reflect on which model is best suited based on the skills required and model performance data above.

3. Route to that model using <search> tags and provide final answer in <answer>...</answer>

Every time you receive new information, you must first conduct reasoning inside <think> ... </think>. After reasoning, if you find you lack some knowledge, you can call a specialized LLM by writing a query inside <search> LLM-Name:Your-Query </search>.
!!! STRICT FORMAT RULES for <search>: !!!
    + You MUST replace LLM-Name with the EXACT name of a model selected from
      [Qwen2.5-7B-Instruct, LLaMA-3.1-8B-Instruct, LLaMA-3.1-70B-Instruct,
       Mistral-7B-Instruct, Mixtral-8x22B-Instruct, Gemma-2-27B-Instruct].
    + You MUST replace Your-Query with the EXACT same question as the original question below (DO NOT CHANGE IT).
    + NEVER copy or paste model descriptions into <search>.
    + NEVER output the placeholder format <search> LLM-Name:Your-Query </search>. Always replace both parts correctly.

Before each LLM call, you MUST explicitly reason inside <think> ... </think> about:
    + Why external information is needed.
    + Which skills from the catalog are required for this question.
    + Which model is best suited based on the model performance data above.

When you call an LLM, the response will be returned between <information> and </information>.
You are encouraged to explore and utilize different LLMs to better understand their respective strengths and weaknesses.

If you find that no further external knowledge is needed, you can directly provide your final answer to the original question inside <answer> ... </answer>, without additional explanation or illustration.

If you think none of the models listed have the necessary skills to answer this question directly, you can route to the model with the highest overall pass rate of models in the pool to get more information.

Question: How many complex numbers satisfy the equation $z^5=\overline{z}$, where $\overline{z}$ is the conjugate of the complex number $z$?

(*@\tagpill{NavyBlue}{Turn 1 — skill analysis + first external call}@*)
<skill_analysis>
{
  "required_skills": [
    {"skill_id": "algebra.logs_and_exponents", "percentage": 70},
    {"skill_id": "algebra.rational_linear_manipulation", "percentage": 30}
  ],
  "reasoning": "Use polar form; equate magnitudes and angles; solve exponent/angle constraints."
}
</skill_analysis>
<think>
To solve this problem, we need to manipulate the given equation involving complex numbers and their conjugates. This requires skills in algebra, specifically dealing with logarithms and exponents, as well as rational/linear manipulation.
</think>
<search>
Mixtral-8x22B-Instruct: How many complex numbers satisfy the equation $z^5=\overline{z}$, where $\overline{z}$ is the conjugate of the complex number $z$?
</search>

<information>
The equation $z^5=\\overline{z}$ implies that $|z^5|=|\\overline{z}|$, which simplifies to $|z|^5=|z|$. 
This implies that $|z|=0$ or $|z|=1$.
- If $|z|=0$, then $z=0$.
- If $|z|=1$, then $z=e^{i\\theta}$ for some $\\theta\\in\\mathbb{R}$. 
Substituting this into the original equation, we get $e^{5i\\theta}=e^{-i\\theta}$. 
This implies that $5\\theta=-\\theta+2k\\pi$ for some $k\\in\\mathbb{Z}$. 
Solving for $\\theta$, we get $\\theta=\\frac{k\\pi}{3}$ for some $k\\in\\mathbb{Z}$.
Therefore, there are infinitely many complex numbers that satisfy the equation $z^5=\\overline{z}$. These are $z=0$ and $z=e^{i\\frac{k\\pi}{3}}$ for any $k\\in\\mathbb{Z}$.
</information>

(*@\tagpill{NavyBlue}{Turn 2 — second call to resolve counting}@*)
<skill_analysis>
{
  "required_skills": [
    {"skill_id": "algebra.logs_and_exponents", "percentage": 60},
    {"skill_id": "algebra.rational_linear_manipulation", "percentage": 40}
  ],
  "reasoning": "Need correct restriction on k from \theta in [0,2\pi)."
}
</skill_analysis>
<think>
We must count distinct solutions modulo 2\pi; query a higher-capability model to confirm.
</think>
<search>
LLaMA-3.1-70B-Instruct: How many complex numbers satisfy the equation $z^5=\overline{z}$, where $\overline{z}$ is the conjugate of the complex number $z$?
</search>

<information>
[...]
Substituting $z = re^{i\theta}$ into the equation $z^5=\overline{z}$, we get
$(re^{i\theta})^5 = re^{-i\theta}$, i.e., $r^5e^{5i\theta} = re^{-i\theta}$.
Equating magnitudes gives $r^5 = r$, so $r = 0$ or $r = 1$.
Equating arguments gives $5\theta = -\theta + 2k\pi$, so $6\theta = 2k\pi$ and
$\theta = \frac{k\pi}{3}$.
Since $\theta \in [0,2\pi)$, we have $0 \le \frac{k\pi}{3} < 2\pi$, so $k=0,1,2,3,4,5$.
For $r = 0$, there is exactly one solution ($z = 0$). 
For $r = 1$, there are 6 possible values of $\\theta$ corresponding to the 6 possible values of $k$. Therefore, the total number of complex numbers that satisfy the equation is 1 (for $r = 0$) + 6 (for $r = 1$) = 7.
</information>

(*@\tagpill{NavyBlue}{Turn 3 — final answer}@*)
<answer>7</answer>

\end{lstlisting}
}

\end{tcolorbox}

\captionof{figure}{Full trace of an AMC example. The router first analyzes the required skills, calls Mixtral-8x22B-Instruct, which derives the correct form but miscounts the solutions; it then routes to LLaMA-3.1-70B-Instruct to correctly restrict $\theta\in[0,2\pi)$ and count distinct solutions, producing the correct final answer.}
\label{fig:example_multi_amc}

\lstset{
  basicstyle=\ttfamily\scriptsize,
  breaklines=true,
  columns=fullflexible,
  keepspaces=true,
  showstringspaces=false,
  aboveskip=0pt,
  belowskip=0pt,
  xleftmargin=0pt,
  frame=none
}

\begin{tcolorbox}[
  title={AMC Example: Skill-Based Router Recognizes No External Model Is Needed and Answers Autonomously},
  sharp corners,
  breakable,
  colframe=Periwinkle,
  colback=white,
  boxrule=2.2pt,
  boxsep=0.9pt,
  enhanced,
  shadow={3pt}{-3pt}{0pt}{opacity=1,mygrey},
]
% --- Key properties strip (move ABOVE listing, not below) ---
\noindent
\statuspill{TagGreen}{\ding{51}\ Skill Handbook}\hspace{0.55em}
\statuspill{TagOrange}{0 external model calls}\hspace{0.6em}
\statuspill{TagGreen}{\ding{51}\ Correct}
\vspace{0.3em}
{\scriptsize
\begin{lstlisting}[language={}, escapeinside={(*@}{@*)}]

(*@\tagpill{cyan}{Skill Router Instruction}@*)
You are a skill-based model router. You are selecting the best model to answer a question by analyzing a question to identify required skills and their importance related to this question.

## Learned Skill Definitions (from validation)

### Algebraic Modeling and Equation Solving
Setting up and solving algebraic relationships, including polynomial constraints, radical systems, and vector-rate models.
  - Polynomial interpolation from structural constraints: Use given values and leading coefficients to determine polynomial constants or invariant sums (e.g., P(0)+Q(0)) without full reconstruction.
    Examples: Problem 1: Quadratic polynomials P and Q with opposite leading coefficients both pass through two points; find P(0)+Q(0).
  - Linearization of symmetric radical systems: Transform symmetric systems with square roots via substitutions, factoring, and controlled squaring to solve for variables or symmetric expressions.
    Examples: Problem 7: Solve a 3-equation symmetric radical system and compute [(1-x)(1-y)(1-z)]^2.
  - Vector kinematics in flowing media: Model motion with currents/winds by decomposing velocities into components relative to ground and medium; impose geometric/temporal constraints to solve for unknowns.
    Examples: Problem 11: Two swimmers in a flowing river head to a common point on the opposite bank; determine the downstream offset D.

### Euclidean Geometry and Transformations
Reasoning about plane and spatial geometry using angle bisectors, tangencies, homothety/inversion, and 3D-to-2D section relations.
  - Circle tangency via homothety/inversion: Exploit homothety or inversion to relate radii, centers, and intersection loci of tangent circles and inscribed figures.
    Examples: Problem 14: Three equal tangent circles inside a circumcircle around an equilateral triangle; find side of the inner equilateral from circle intersections.
  - Angle bisectors and (ex)center configurations: Use properties of angle bisectors, incenter/excenters, and their intersections to compute distances in polygons, especially trapezoids and triangles.
    Examples: Problem 9: In an isosceles trapezoid, angle bisectors meet at P and Q; find PQ given side lengths.
  - Sphere-plane sections and power-of-a-point relations: Relate congruent circular cross-sections of spheres to sphere radii and plane offsets; compute distances between projected centers using tangency and power.
    Examples: Problem 2: Three tangent spheres cut by a plane in congruent circles with centers A,B,C; given AB^2, find AC^2.
  - Equal-perimeter splitting lines via reflection: Model equal-perimeter partitions by reflecting across sides to convert perimeter conditions to straight-line distance constraints.
    Examples: Problem 6: Splitting lines through midpoints M and N in a triangle intersect; deduce side lengths/structure from equal-perimeter property.

### Combinatorics and Discrete Structures
Counting and optimization in discrete settings with indicator methods, parity constraints, pattern avoidance, and digit-product arrangements.
  - Indicator variables and linearity over subsets: Compute sums over families of sets by summing elementwise contributions with indicator variables and symmetry constraints like |A|=|B|.
    Examples: Problem 4: Evaluate $S_n = \sum |A \cap B|$ over ordered pairs of subsets with equal cardinalities.
  - AP-free (pattern-avoidance) sequence design: Select or count integers under monotonicity while forbidding k-term arithmetic progressions; use modular classes and structural constraints.
    Examples: Problem 12: Count integer pairs (a,b) so the 8-term increasing sequence contains no 4-term arithmetic progression.
  - Parity-constrained arrangements with identical pairs: Count permutations where identical items must be separated by an even/odd number of positions; apply parity classes and inclusion-exclusion or structural bijections.
    Examples: Problem 15: Probability a random arrangement of two each of six colors is 'even' (parity gap constraint).
  - Digit-based discrete optimization of product ratios: Assign distinct digits to maximize/minimize product differences/ratios under constraints using inequalities, rearrangement, and greedy swaps.
    Examples: Problem 13: Minimize a ratio of product difference over product using digits 1-9 exactly once; report m+n.

### Number Theory and Modular Reasoning
Reasoning with modular arithmetic, base representations, reduced fractions of repeating decimals, and roots-of-unity congruences.
  - Repeating decimal reduction via gcd/phi structure: Convert repeating decimals to fractions with denominators of the form 9, 99, 9999..., reduce by gcd, and count distinct reduced numerators leveraging multiplicative structure.
    Examples: Problem 5: Count distinct numerators obtained from all 4-digit repeating decimals when reduced to lowest terms.
  - Base conversion as digit-constrained Diophantine equations: Translate cross-base digit identities into linear equations with digit bounds; solve for digits satisfying both base systems.
    Examples: Problem 8: Find three-digit base-10 abc whose base-9 representation is bca_9.
- Roots-of-unity angle congruences and counting: Interpret complex constants as $e^{i\theta}$; equate powers via modular congruences on arguments and count solutions within given bounds.
  Examples: Problem 10: With $w = \frac{\sqrt{3}+i}{2}$ and $z = \frac{-1+i\sqrt{3}}{2}$, count $(r,s) \le 100$ satisfying $i w^r = z^s$.

## Model Performance (learned from validation)

### LLaMA-3.1-70B-Instruct
Overall: 30% success (6/20)
Skill scores:
  - Sphere-plane sections and power-of-a-point relations: 100%
  - Base conversion as digit-constrained Diophantine equations: 100%
  - Parity-constrained arrangements with identical pairs: 100%
  - Circle tangency via homothety/inversion: 50%
  - Indicator variables and linearity over subsets: 50%
Strengths: 3D geometry with spheres/planes and Power of a Point, Number base representation and digit-constraint puzzles, Best overall reliability in this batch
Weaknesses: Circle tangency/homothety/inversion geometry, Algebra with polynomial interpolation/constraint reasoning, Subset-sum/inclusion-exclusion style combinatorics

### Mistral-7B-Instruct
Overall: 5% success (1/20)
Skill scores:
  - Indicator variables and linearity over subsets: 50%
  - Polynomial interpolation from structural constraints: 0%
  - Sphere-plane sections and power-of-a-point relations: 0%
  - Circle tangency via homothety/inversion: 0%
  - Angle bisectors and (ex)center configurations: 0%
Strengths: Occasional general reasoning success (inconsistent)
Weaknesses: Euclidean geometry (especially tangency/homothety/inversion), 3D geometry with spheres/planes, Polynomial interpolation constraints

### Gemma-2-27B-Instruct
Overall: 0% success (0/20)
Skill scores:
  - Polynomial interpolation from structural constraints: 0%
  - Sphere-plane sections and power-of-a-point relations: 0%
  - Circle tangency via homothety/inversion: 0%
  - Angle bisectors and (ex)center configurations: 0%
  - Indicator variables and linearity over subsets: 0%
Weaknesses: 3D sphere-plane geometry, Circle tangency/homothety/inversion, Polynomial interpolation constraints

### Mixtral-8x22B-Instruct
Overall: 0% success (0/20)
Skill scores:
  - Polynomial interpolation from structural constraints: 0%
  - Sphere-plane sections and power-of-a-point relations: 0%
  - Circle tangency via homothety/inversion: 0%
  - Angle bisectors and (ex)center configurations: 0%
  - Indicator variables and linearity over subsets: 0%
Weaknesses: 3D sphere-plane geometry, Circle tangency/homothety/inversion, Polynomial interpolation constraints

### Qwen2.5-7B-Instruct
Overall: 0% success (0/20)
Skill scores:
  - Polynomial interpolation from structural constraints: 0%
  - Sphere-plane sections and power-of-a-point relations: 0%
  - Circle tangency via homothety/inversion: 0%
  - Angle bisectors and (ex)center configurations: 0%
  - Indicator variables and linearity over subsets: 0%
Weaknesses: Geometry-heavy tasks, Algebraic modeling/constraints, 0% on tested tasks

### LLaMA-3.1-8B-Instruct
Overall: 0% success (0/20)
Skill scores:
  - Polynomial interpolation from structural constraints: 0%
  - Sphere-plane sections and power-of-a-point relations: 0%
  - Circle tangency via homothety/inversion: 0%
  - Angle bisectors and (ex)center configurations: 0%
  - Indicator variables and linearity over subsets: 0%
Weaknesses: All observed olympiad-style math skills in this batch, All tested math skills in this batch (0% accuracy)

## Cost Tiers (cheapest to most expensive)
- Cheap: Qwen2.5-7B-Instruct, LLaMA-3.1-8B-Instruct, Mistral-7B-Instruct
- Medium: Gemma-2-27B-Instruct
- Expensive: LLaMA-3.1-70B-Instruct, Mixtral-8x22B-Instruct

## Task
1. First, analyze the question below and identify which skills are needed, along with the percentage/weight of each skill (how important each skill is for answering this question).

Output your analysis in the following format inside <skill_analysis> tags:
<skill_analysis>
{
  "required_skills": [
    {"skill_id": "skill.id", "percentage": 50},
    {"skill_id": "skill.id", "percentage": 30},
    ...
  ],
  "reasoning": "Brief explanation of why these skills are needed"
}
</skill_analysis>

The percentages should sum to approximately 100 (they don't need to be exact, but should reflect relative importance).

2. After providing the skill analysis, reflect on which model is best suited based on the skills required and model performance data above.

3. Route to that model using <search> tags and provide final answer in <answer>...</answer>

Every time you receive new information, you must first conduct reasoning inside <think> ... </think>. After reasoning, if you find you lack some knowledge, you can call a specialized LLM by writing a query inside <search> LLM-Name:Your-Query </search>. 
!!! STRICT FORMAT RULES for <search>: !!!
    + You MUST replace LLM-Name with the EXACT name of a model selected from [Qwen2.5-7B-Instruct, LLaMA-3.1-8B-Instruct, LLaMA-3.1-70B-Instruct, Mistral-7B-Instruct, Mixtral-8x22B-Instruct, Gemma-2-27B-Instruct].     
    + You MUST replace Your-Query with the EXACT same question as the original question below (DO NOT CHANGE IT).     
    + NEVER copy or paste model descriptions into <search>.
    + NEVER output the placeholder format <search> LLM-Name:Your-Query </search>. Always replace both parts correctly. 
Before each LLM call, you MUST explicitly reason inside <think> ... </think> about:     
    + Why external information is needed.     
    + Which skills from the catalog are required for this question.     
    + Which model is best suited based on the model performance data above. 
When you call an LLM, the response will be returned between <information> and </information>. You are encouraged to explore and utilize different LLMs to better understand their respective strengths and weaknesses. 
If you find that no further external knowledge is needed, you can directly provide your final answer to the original question inside <answer> ... </answer>, without additional explanation or illustration. For example: <answer> Beijing </answer>.     
+ Important: You must not output the placeholder text "<answer> and </answer>" alone.     
+ You must insert your actual answer between <answer> and </answer>, following the correct format.   
+ You must not output the model name or query between <answer> and </answer>. 
If you think none of the models listed have the necessary skills to answer this question directly, you can route to the model with the highest overall pass rate of models in the pool to get more information.

Question: A $3-4-5$ right triangle is inscribed in circle $A$, and a $5-12-13$ right triangle is inscribed in circle $B$. Find the ratio of the area of circle $A$ to the area of circle $B$. The final answer can be written in the form $\frac{m}{n}$, where $m$ and $n$ are relatively prime positive integers. What is $m+n$?

(*@\tagpill{NavyBlue}{Turn 1 - direct answer}@*)
<think>
To solve this problem, we need to find the radii of the circles in which the right triangles are inscribed. For a right triangle, the hypotenuse is the diameter of the circumscribed circle. Therefore, we can find the radius of each circle by dividing the length of the hypotenuse by 2.

For the $3-4-5$ right triangle, the hypotenuse is 5, so the radius of circle $A$ is $\frac{5}{2}$. For the $5-12-13$ right triangle, the hypotenuse is 13, so the radius of circle $B$ is $\frac{13}{2}$.

The area of a circle is given by $\pi r^2$. Therefore, the area of circle $A$ is $\pi \left(\frac{5}{2}\right)^2 = \pi \frac{25}{4}$, and the area of circle $B$ is $\pi \left(\frac{13}{2}\right)^2 = \pi \frac{169}{4}$.

The ratio of the area of circle $A$ to the area of circle $B$ is:
\[
\frac{\pi \frac{25}{4}}{\pi \frac{169}{4}} = \frac{25}{169}
\]
Since 25 and 169 are relatively prime, the ratio is $\frac{25}{169}$. Therefore, $m+n = 25+169 = 194$.

No further information is needed, so I can directly provide the answer.
</think>
<answer>194</answer>

\end{lstlisting}
}
\end{tcolorbox}
\captionof{figure}{Full trace of an AMC example. While the router (Qwen2.5-3B) has access to the Skill Handbook, it autonomously determines that the problem can be solved using its internal capability, refrains from issuing any \texttt{<search>} calls, and produces the correct final answer. This example highlights the flexibility of skill-based routing, where external model invocation is optional rather than mandatory.}
\label{fig:trace_no_help_amc}

\begin{tcolorbox}[
  title={PopQA Example: Adaptive Skill-Based Routing Recovers from a Noisy Initial Response},
  sharp corners,
  breakable,
  colframe=Periwinkle,
  colback=white,
  boxrule=2.2pt,
  boxsep=0.9pt,
  enhanced,
  shadow={3pt}{-3pt}{0pt}{opacity=1,mygrey},
]
\noindent
\statuspill{TagGreen}{\ding{51}\ Skill Handbook}\hspace{0.6em}
\statuspill{TagOrange}{3 external calls}\hspace{0.6em}
\statuspill{TagGreen}{\ding{51}\ Correct}
\vspace{0.35em}

{\scriptsize
\begin{lstlisting}[language={}, escapeinside={(*@}{@*)}]
(*@\tagpill{cyan}{Skill Router Instruction}@*)
You are a skill-based model router. You are selecting the best model to answer a question by analyzing a question to identify required skills and their importance related to this question.

## Learned Skill Definitions (from validation)

### Entity Resolution and Attributes
Identify and link entities from descriptive phrases and retrieve their attributes or locations.
  - Descriptor-based Entity Linking: Resolve a specific entity from multi-clause descriptors (titles, awards, roles) without the name explicitly given.
    Examples: Problem 12: Who is the American singer-songwriter, who won an award for Best Female Video at the 2009 MTV Video Music Awards, and wrote a song for the 'AT&T Team USA Soundtrack'?, Problem 1: Who is the English Professional Footballer that is the niece of a former footballer born in the same place as where he plays?
  - Membership/Role Resolution: Identify an entity based on group membership and an additional role or credit.
    Examples: Problem 4: What member of the South Korean-Chinese boy group EXO stars in the upcoming movie "The Underdog"?, Problem 9: Maha Kali is an EP by what band from Stromstad that was formed in 1989?
  - Attribute/Location Retrieval: Extract a specific attribute or location associated with an entity from knowledge bases.
    Examples: Problem 15: Japanese Weekend School of New York has offices in the entertainment complex in what section of New Rochelle?

### Relational Composition and Constraints
Answer questions requiring chaining multiple relations and satisfying constraints across entities.
  - Multi-hop Bridge Reasoning: Combine two or more linked facts (A->B->C) to derive the answer.
    Examples: Problem 6: The Battle of Cambrai took place during a series of offensives that ended on what date?, Problem 3: Liz Rose has co-written songs with which artist including "White Horse" and "You Belong with Me"?
  - Relational Constraint Satisfaction: Apply constraints such as shared attributes (same birthplace/place) or time/place filters across relations to identify the correct entity.
    Examples: Problem 1: Who is the English Professional Footballer that is the niece of a former footballer born in the same place as where he plays?, Problem 9: Maha Kali is an EP by what band from Stromstad that was formed in 1989?
  - Reverse Relation Traversal: Start from works or properties and infer the originating entity via reverse edges (e.g., song -> artist).
    Examples: Problem 3: Liz Rose has co-written songs with which artist including "White Horse" and "You Belong with Me"?, Problem 10: A Pair of Brown Eyes and Wild Mountain Thyme is based from what artists song?

### Temporal and Ordinal Reasoning
Interpret dates, periods, and ordinal descriptors to resolve time-based queries.
  - Event-Date Alignment: Map events to their dates directly or via higher-level campaigns/series.
    Examples: Problem 6: The Battle of Cambrai took place during a series of offensives that ended on what date?
  - Ordinal Title/Sequence Disambiguation: Use ordinal descriptors (e.g., third, first) within a known sequence to identify the correct entity and associated dates.
    Examples: Problem 5: When did the the Antonine who was known as the third of the five good emperors live?
  - Temporal Qualifier Interpretation: Interpret temporal qualifiers such as 'first', 'upcoming', or year references to locate the relevant time frame.
    Examples: Problem 4: What member of the South Korean-Chinese boy group EXO stars in the upcoming movie "The Underdog"?, Problem 8: Shirley Breeden won her first Senate term in a narrow upset over the politician who was a member of what branch of the armed forces?

### Comparative, Classification, and Set Reasoning
Make comparisons, classify options by properties, and find common attributes across entities.
  - Property-based Option Selection: Choose among given options based on a specific property filter.
    Examples: Problem 2: Which is a black-and-white film, Flying Padre or Inside Job?
  - Numeric/Quantitative Comparison: Compare numeric attributes (counts, totals) across entities to determine the greater/lesser.
    Examples: Problem 14: Which player won more Grand Slam titles, Kevin Ullyett or Billie Jean King?
  - Set Intersection/Common Attribute Identification: Find a shared attribute (e.g., profession) between two entities.
    Examples: Problem 13: What profession does Leonty Magnitsky and Leonid Khachiyan have in common?

### Language Robustness and Causal Inference
Handle noisy phrasing and infer causal or thematic relations from text.
  - Noisy/Paraphrase Normalization: Parse ungrammatical or awkward phrasing and map it to a coherent structured query.
    Examples: Problem 10: A Pair of Brown Eyes and Wild Mountain Thyme is based from what artists song?, Problem 11: In which song was written by singer-songwriter Taylor Swift and shares the optimistic lyrical message to a song called "Yodel It!"?
  - Causal/Narrative Why-Reasoning: Answer why-questions by identifying causes, motives, or precipitating events in historical narratives.
    Examples: Problem 7: Why did Rudolf Hess stop serving Hitler in 1941?
  - Thematic/Message Similarity Inference: Infer thematic similarity (e.g., optimistic message) between works when not explicitly linked by facts.
    Examples: Problem 11: In which song was written by singer-songwriter Taylor Swift and shares the optimistic lyrical message to a song called "Yodel It!"?

## Model Performance (learned from validation)

### LLaMA-3.1-70B-Instruct
Overall: 43% success (26/60)
Skill scores:
  - Property-based Option Selection: 100%
  - Numeric/Quantitative Comparison: 100%
  - Membership/Role Resolution: 67%
  - Thematic/Message Similarity Inference: 67%
  - Noisy/Paraphrase Normalization: 60%
Strengths: Property-based option selection, Stronger-than-peers multi-hop bridge reasoning, Paraphrase/noise normalization
Weaknesses: Descriptor-based entity linking, Relational constraint satisfaction, Relational constraint satisfaction across multiple entities

### Gemma-2-27B-Instruct
Overall: 48% success (29/60)
Skill scores:
  - Property-based Option Selection: 100%
  - Set Intersection/Common Attribute Identification: 100%
  - Numeric/Quantitative Comparison: 100%
  - Thematic/Message Similarity Inference: 67%
  - Attribute/Location Retrieval: 50%
Strengths: Property-based option selection (top-tier), Paraphrase/noise normalization, Thematic similarity inference
Weaknesses: Multi-hop bridge reasoning (mid), Descriptor-based entity linking, Relational constraint satisfaction

### Mistral-7B-Instruct
Overall: 30% success (18/60)
Skill scores:
  - Property-based Option Selection: 100%
  - Numeric/Quantitative Comparison: 100%
  - Attribute/Location Retrieval: 39%
  - Reverse Relation Traversal: 30%
  - Multi-hop Bridge Reasoning: 27%
Strengths: Occasional success on simple entity lookups, No demonstrated strengths on measured skills in this set, Cost/latency for very simple, single-hop queries
Weaknesses: Multi-hop bridge reasoning (low), Descriptor-based entity linking, Relational constraint satisfaction

### Mixtral-8x22B-Instruct
Overall: 45% success (27/60)
Skill scores:
  - Property-based Option Selection: 100%
  - Membership/Role Resolution: 100%
  - Set Intersection/Common Attribute Identification: 100%
  - Numeric/Quantitative Comparison: 100%
  - Attribute/Location Retrieval: 44%
Strengths: Multi-hop bridge reasoning (strong among evaluated models), Descriptor-based entity linking (partial strength), Multi-hop bridge reasoning (large-model tier)
Weaknesses: Descriptor-based entity linking, Relational constraint satisfaction, Paraphrase/noise normalization

### LLaMA-3.1-8B-Instruct
Overall: 25% success (15/60)
Skill scores:
  - Property-based Option Selection: 100%
  - Set Intersection/Common Attribute Identification: 67%
  - Numeric/Quantitative Comparison: 50%
  - Event-Date Alignment: 29%
  - Attribute/Location Retrieval: 22%
Strengths: Basic multi-hop (limited), No demonstrated strengths on measured skills in this set, Basic single-hop factoids
Weaknesses: Multi-hop bridge reasoning (below larger models), Descriptor-based entity linking, Relational constraint satisfaction

### Qwen2.5-7B-Instruct
Overall: 33% success (20/60)
Skill scores:
  - Property-based Option Selection: 100%
  - Set Intersection/Common Attribute Identification: 100%
  - Numeric/Quantitative Comparison: 100%
  - Attribute/Location Retrieval: 56%
  - Multi-hop Bridge Reasoning: 21%
Strengths: None clearly demonstrated in this sample, Partial capability in paraphrase/noise normalization, Simple fact lookup with low latency
Weaknesses: Multi-hop bridge reasoning (0%), Descriptor-based entity linking, Relational constraint satisfaction

## Cost Tiers (cheapest to most expensive)
- Cheap: Qwen2.5-7B-Instruct, LLaMA-3.1-8B-Instruct, Mistral-7B-Instruct
- Medium: Gemma-2-27B-Instruct
- Expensive: LLaMA-3.1-70B-Instruct, Mixtral-8x22B-Instruct

## Task
1. First, analyze the question below and identify which skills are needed, along with the percentage/weight of each skill (how important each skill is for answering this question).

Output your analysis in the following format inside <skill_analysis> tags:
<skill_analysis>
{
  "required_skills": [
    {"skill_id": "skill.id", "percentage": 50},
    {"skill_id": "skill.id", "percentage": 30},
    ...
  ],
  "reasoning": "Brief explanation of why these skills are needed"
}
</skill_analysis>

2. After providing the skill analysis, reflect on which model is best suited based on the skills required and model performance data above.

3. Route to that model using <search> tags and provide final answer in <answer>...</answer>

Every time you receive new information, you must first conduct reasoning inside <think> ... </think>. After reasoning, if you find you lack some knowledge, you can call a specialized LLM by writing a query inside <search> LLM-Name:Your-Query </search>.
!!! STRICT FORMAT RULES for <search>: !!!
    + You MUST replace LLM-Name with the EXACT name of a model selected from
      [Qwen2.5-7B-Instruct, LLaMA-3.1-8B-Instruct, LLaMA-3.1-70B-Instruct, Mistral-7B-Instruct,
       Mixtral-8x22B-Instruct, Gemma-2-27B-Instruct].
    + You MUST replace Your-Query with the EXACT same question as the original question below (DO NOT CHANGE IT).
    + NEVER copy or paste model descriptions into <search>.
    + NEVER output the placeholder format <search> LLM-Name:Your-Query </search>. Always replace both parts correctly.
Before each LLM call, you MUST explicitly reason inside <think> ... </think> about:
    + Why external information is needed.
    + Which skills from the catalog are required for this question.
    + Which model is best suited based on the model performance data above.
When you call an LLM, the response will be returned between <information> and </information>.
If you find that no further external knowledge is needed, you can directly provide your final answer to the original question inside <answer> ... </answer>, without additional explanation.

Question: Who was the composer of It's Going to Take Some Time?

(*@\tagpill{NavyBlue}{Turn 1 — noisy lookup}@*)
<think>
To answer this question, we need to identify the composer of the song \"It's Going to Take Some Time\". This requires the skill of Reverse Relation Traversal, which involves starting from a work (in this case, the song) and inferring the originating entity (the composer). We also need the skill of Attribute/Location Retrieval, as we are looking for a specific attribute (composer) associated with the song.
</think>

<skill_analysis>
{
  "required_skills": [
    {"skill_id": "Reverse Relation Traversal", "percentage": 60},
    {"skill_id": "Attribute/Location Retrieval", "percentage": 40}
  ],
  "reasoning": "We need to start from the song and infer the composer, and then retrieve the composer's attribute."
}
</skill_analysis>

<search>Qwen2.5-7B-Instruct: Who was the composer of It's Going to Take Some Time?</search>

<information>
The song "It's Going to Take Some Time" is a popular track from the musical "The Wedding Singer."
The composer of this song is Frank Wildhorn. Frank Wildhorn is an American composer known for his work in musical theater, including other notable works such as "Jekyll & Hyde" and "The Last 5 Years."
He collaborated with lyricist Michael Gore on this particular song. The song was featured in the 1998 film adaptation of "The Wedding Singer," which starred Adam Sandler.
Would you like more information about Frank Wildhorn or the musical "The Wedding Singer"? If not, I can provide insights on the song's significance in the context of the musical. Let me know your next step.
[... noisy repetitive content omitted: the same "let me know" lines repeat many times ...]
</information>

(*@\tagpill{NavyBlue}{Turn 2 - uncertain attribution}@*)
<think>
The Qwen2.5-7B-Instruct model couldn't provide a definitive answer and includes excessive filler content. I need to consult another model with stronger factual knowledge about music.
</think>

<search>LLaMA-3.1-70B-Instruct: Who was the composer of It's Going to Take Some Time?</search>

<information>
"It's Going to Take Some Time" is a song written by Toni Stern and Carole King. It was recorded by Carole King in 1971.
However, I am not sure if she was the composer of the song.
</information>

(*@\tagpill{NavyBlue}{Turn 3 — cross verification}@*)
<think>
LLaMA-3.1-70B identifies Toni Stern and Carole King as the writers, but explicitly expresses uncertainty about whether Carole King should be treated as the "composer" requested. Cross-verify with another model.
</think>

<search>Mixtral-8x22B-Instruct: Who was the composer of It's Going to Take Some Time?</search>

<information>
The composer of "It's Going to Take Some Time" is Carole King. She co-wrote the song with Toni Stern.
The song was first released on her 1971 album "Music" and later became a hit single for The Carpenters in 1972.
</information>
(*@\tagpill{NavyBlue}{Turn 4 — final answer}@*)
<answer>Carole King</answer>
\end{lstlisting}
}
\end{tcolorbox}

\captionof{figure}{Full trace for a PopQA example. Guided by skill-based analysis, the router detects that the initial model response is incomplete or ambiguous, re-routes to alternative models, and performs cross-model verification before finalizing the answer. This demonstrates robust recovery from intermediate errors.}
\label{fig:router_trace_popqa_noisy}

\begin{tcolorbox}[
  title={Agent Orchestration Instruction (Full Specification)},
  sharp corners,
  breakable,
  colframe=RoyalBlue,
  colback=white,
  boxrule=2.2pt,
  boxsep=0.9pt,
  enhanced,
  shadow={3pt}{-3pt}{0pt}{opacity=1,mygrey},
]

\begin{lstlisting}[language={}, basicstyle=\ttfamily\scriptsize]
You are a skill-based orchestrator for multi-step question answering. 
You select the best tool (search|code|answer) and model by analyzing required skills.
\end{lstlisting}

\vspace{0.5em}
\begin{lstlisting}[language={}, basicstyle=\ttfamily\scriptsize]
## Problem to Solve
{{problem}}
\end{lstlisting}

\vspace{0.5em}

\begin{lstlisting}[language={}, basicstyle=\ttfamily\scriptsize]
## Current Context
{{context}}

python code and execution outputs:
{{execution_results}}
\end{lstlisting}

\medskip
\hrule
\medskip

\begin{lstlisting}[language={}, basicstyle=\ttfamily\scriptsize]
## Quick Reference: What You Need to Do

CRITICAL REQUIREMENT: Before making ANY tool call, you MUST:

1. Inside <think>...</think>, analyze required skills and output in <skill_analysis> tags.
2. Then choose the appropriate tool with the selected model based on the skill analysis.

The context above may be long - scroll back to see the problem and context, then follow the instructions below.
\end{lstlisting}

\medskip
\hrule
\medskip

\begin{lstlisting}[language={}, basicstyle=\ttfamily\scriptsize]
## Available Tools and Models (Baseline)

Tool: search
Models: search-1 ($10/M), search-2 ($2/M), search-3 ($0.8/M)
Description: Search for missing information

Tool: code | enhance_reasoning
Models: reasoner-1 ($10/M), reasoner-2 ($2/M), reasoner-3 ($0.8/M)
Description: Write and execute Python code to solve the problem

Tool: answer
Models: answer-1 ($10/M), answer-2 ($2/M), answer-3 ($0.9/M),
        answer-4 ($0.8/M), answer-math-1 ($0.9/M), answer-math-2 ($0.2/M)
Description: Extract the final answer if you think you have enough information to answer the problem
\end{lstlisting}

\medskip
\hrule
\medskip

\begin{lstlisting}[language={}, basicstyle=\ttfamily\scriptsize]
## Learned Skill Definitions

### Search Skills

- search.control.zero_result_guardrail: Detect zero-result searches and prevent escalation to non-search stages until several structured reformulations and decomposition attempts have been tried.
- search.control.stage_adherence: Adhere to the search stage's intended tools and behaviors; avoid misusing the code tool when the task requires query generation and retrieval.
- search.control.answer_block_on_zero_results: Prevent advancing to the answer stage when searches return zero results; require iterative reformulation attempts first.
- search.query_planning.multi-hop.list_then_filter.attribute_match: Plan retrieval by first listing entities that satisfy a temporal/role constraint, then filtering them by a specific attribute or name match (e.g., middle name matching a category like Scottish monarchs).
- search.tool_choice.scope_awareness: Avoid search for well-known facts that can be confidently recalled; reserve search for low-confidence or obscure queries.
- search.efficiency.tool_call_minimization: Avoids unnecessary tool invocations (search or code) when they do not contribute to retrieval, reducing cost and noise.
- search.tool_choice.code_for_computation_only: Uses the code tool solely for numerical or data-processing steps after retrieval, not for information retrieval or scraping.
- search.retrieval.robust_authoritative_search: Plan and execute single- and multi-hop web retrieval with entity/event disambiguation and numeric/temporal reasoning while explicitly prioritizing authoritative sources (e.g., Britannica/Wikipedia for general facts, IOC/FIFA/World Athletics for sports, NOAA/NASA and peer-reviewed literature for science, official corporate and regulator sites for filings). Construct resilient queries using operators and filters (quotes, OR, -, site:, filetype:, language, date ranges); align dates, ranges, and units; iteratively broaden/narrow via synonyms and result feedback; detect empty/malformed or low-quality-result queries and systematically retry/refine (including minimal viable queries) until at least one high-quality supporting source is obtained or escalate for clarification if recovery fails.
Examples: Compute 10 years after Neil Armstrong's birth, then search nasa.gov with date filters for events on that date; refine queries if initial filters yield no results., Who won the women's marathon at the 2024 Olympics? Prioritize olympics.com or World Athletics results; apply year filters and exact-match athlete names; retry with synonyms or broader terms if initial query is sparse.
- search.strategy.query_formulation_refinement_and_verification: Formulate clear, targeted search queries using key entities and relations; when results are weak or conflicting, rapidly iterate with synonyms, paraphrases, broader/narrower terms, constraints, operators, time filters, and source scoping. Perform lightweight verification searches even when an answer is recalled, corroborate with credible sources, and adjust conclusions based on evidence.
Examples: Need: When was the Kyoto Protocol adopted? Start with 'Kyoto Protocol adopted date'; iterate with 'Kyoto Protocol adoption 1997', add site:unfccc.int and time filters to confirm., Recall an author of a concept; perform a brief search to confirm the name and cite the seminal paper, revising if results conflict.
# Additional skills with performance data available:
- retrieval.date_lookup
- retrieval.entity_lookup
- retrieval.multi_hop
- search.basic_retrieval
- search.control.stage_adherence.strict
- search.cost_effectiveness
- search.cost_efficiency
- search.date_lookup
- search.disambiguation
- search.entity_lookup
- search.entity_retrieval
- search.fact_retrieval
- search.factual.lookup
- search.factual_single_hop
- search.lookup.dates
- search.lookup.entities
- search.multi_hop
- search.multi_hop.light
- search.multi_hop.simple
- search.multi_hop_facts
- search.multi_hop_light
- search.multihop
- search.precision.general
- search.query_planning.multi-hop
- search.recall.general
- search.retrieval.cost_efficiency
- search.retrieval.dates
- search.retrieval.factoid
- search.retrieval.facts
- search.retrieval.multi_hop
- search.retrieval.multi_hop_light
- search.retrieval.precision
- search.retrieval.recall
- search.retrieval.single_fact
- search.retrieval.single_hop
- search.single_hop
- search.strategy.source_prioritization
- search.web.fact_lookup
- search.web.lookup
- search.web.multi_hop
\end{lstlisting}

\vspace{0.5em}

\begin{lstlisting}[language={}, basicstyle=\ttfamily\scriptsize]
## Reasoning Skills

- code.generation.codeblock_integrity: Maintain clean code blocks containing only valid Python, avoiding mixed prose or unrelated text.
- code.math.date_calculation: Computing differences between dates, handling month/day boundaries and leap years
- code.data.aggregation: Summing, averaging, grouping, and comparing values across lists or tables
- code.math.multi_step: Accurate computation across multiple operations or items where mental math is error-prone.\n- code.execution.iteration_budgeting: Limits repeated tool invocations; detects diminishing returns and switches strategy (e.g., back to search/answer) to avoid wasted cost.
- code.execution.token_economy: Generates concise code and avoids unnecessary large literals or pasted text to minimize token usage and cost.
- code.stage_adherence: Ensure the code stage produces valid, executable Python only; avoid narrative text or non-code dumps.
- code.execution.strategy_handoff_on_failure: After detecting that code execution cannot progress (e.g., needs external data or repeated non-executable outputs), emit a clear handoff signal to search/answer instead of retrying.
- code.data.tabular_parsing_ops_ranking: Extract structured tables from semi-structured text (lists, textual/markdown/CSV-like tables) and perform tabular filtering, multi-key sorting, stable tie-aware ranking (top/bottom N with ties), grouping, and aggregation. Supports producing rank annotations and operating on small tables or tables returned by search.
Examples: Parse this markdown table of product sales, then return the top 10 products by revenue (include ties) and compute average revenue per category., From this bullet list of shipped orders, extract fields (order_id, customer, ship_date, total), filter status='shipped', sort by ship_date descending, then group by customer and sum total_spent.
- code.execution.offline_minimal_inline_structures: Decide whether code execution is warranted for a task. When appropriate, produce the smallest self-contained, offline-safe Python snippet. Replace external I/O with tiny in-memory structures (lists/dicts/DataFrames) to perform lookups and computations. Validate outputs against requested type/format/schema/units and iterate until compliant. When code adds no value or violates constraints, abstain and answer directly while avoiding noisy outputs.
Examples: A request to map a few codes to names without using files or APIs; construct a small inline dict and return the mapped results., Summarize a CSV hosted at a URL in a no-network environment; create a minimal inline DataFrame with representative rows, compute summary stats offline, and print validated outputs.
- code.datetime.date_difference: Compute accurate differences between dates (years/months/days) with proper inclusivity and leap-year handling, and generate/execute Python datetime code (including parsing) to achieve the same.
Examples: What is the difference between 2020-02-29 and 2021-03-01 in years, months, and days (inclusive)?, Write Python code to parse two dates and return the year, month, and day deltas, correctly handling leap years.
# Additional skills with performance data available:
- code.date.arithmetic
- code.execution.syntax_self_check
- code.math.arithmetic
\end{lstlisting}

\vspace{0.5em}

\begin{lstlisting}[language={}, basicstyle=\ttfamily\scriptsize]
## Answer Skills
No dedicated skills defined.
Use general performance of answer models to select the best model for the answer stage if you think you have enough information to answer the problem.
\end{lstlisting}

\medskip
\hrule
\medskip

\begin{lstlisting}[language={}, basicstyle=\ttfamily\scriptsize]
## Model Performance (learned from validation)

### Search Models

#### search-1 (gpt-5)
Overall: 55% success (11/20)
Skill scores:
  - search.factual_single_hop: 100%
  - search.multi_hop: 100%
  - search.retrieval.multi_hop: 100%
  - search.basic_retrieval: 100%
  - search.retrieval.single_hop: 67%
  - search.tool_choice.scope_awareness: 67%
  - search.factual.lookup: 60%
  - search.fact_retrieval: 60%
  - search.retrieval.cost_efficiency: 50%
  - search.cost_effectiveness: 50%
  - search.cost_efficiency: 40%
  - search.retrieval.precision: 40%
  - search.retrieval.recall: 40%
  - search.control.stage_adherence: 17%
  - search.query_planning.multi-hop: 0%
  - search.control.zero_result_guardrail: 0%
  - search.control.answer_block_on_zero_results: 0%
  - search.control.stage_adherence.strict: 0%
  - search.tool_choice.code_for_computation_only: 0%
  - search.efficiency.tool_call_minimization: 0%
  - search.retrieval.robust_authoritative_search: 0%
  - search.strategy.query_formulation_refinement_and_verification: 0%
Strengths:
  - answered from recall despite zero search results
  - robust multi-hop retrieval
  - maintained stage separation (did not misuse code tool)
Weaknesses:
  - invoked code tool with no effect (unnecessary tool usage)
  - no effective search results and no reformulation
  - issued a search call but returned no results

#### search-2 (gpt-5-mini)
Overall: 50% success (10/20)
Skill scores:
  - search.cost_efficiency: 100%
  - search.basic_retrieval: 100%
  - search.fact_retrieval: 100%
  - search.multi_hop: 67%
  - search.retrieval.single_fact: 67%
  - search.retrieval.precision: 60%
  - search.retrieval.recall: 60%
  - search.retrieval.multi_hop: 60%
  - search.tool_choice.scope_awareness: 50%
  - search.control.stage_adherence: 20%
  - search.query_planning.multi-hop: 0%
  - search.strategy.source_prioritization: 0%
  - search.control.zero_result_guardrail: 0%
  - search.control.answer_block_on_zero_results: 0%
  - search.control.stage_adherence.strict: 0%
  - search.tool_choice.code_for_computation_only: 0%
  - search.efficiency.tool_call_minimization: 0%
  - search.retrieval.robust_authoritative_search: 0%
  - search.strategy.query_formulation_refinement_and_verification: 0%
Strengths:
  - accurate recall of both anchor years and mental subtraction
  - concise final answer
  - maintained stage separation (did not misuse code tool)
Weaknesses:
  - no effective search results
  - no reformulation after zero results
  - relied on an unverified answer without retrieval

#### search-3 (Qwen/Qwen3-32B)
Overall: 65% success (13/20)
Skill scores:
  - search.retrieval.factoid: 100%
  - search.factual_single_hop: 100%
  - search.cost_efficiency: 100%
  - search.date_lookup: 100%
  - search.recall.general: 100%
  - search.precision.general: 100%
  - search.single_hop: 100%
  - search.retrieval.facts: 100%
  - search.retrieval.dates: 100%
  - search.lookup.entities: 100%
  - search.lookup.dates: 100%
  - search.multi_hop.light: 100%
  - search.factual.lookup: 100%
  - search.multi_hop.simple: 100%
  - search.basic_retrieval: 100%
  - search.fact_retrieval: 100%
  - search.disambiguation: 100%
  - search.web.lookup: 100%
  - search.entity_retrieval: 100%
  - search.multi_hop_facts: 100%
  - search.retrieval.single_fact: 100%
  - search.web.fact_lookup: 100%
  - search.web.multi_hop: 100%
  - search.cost_effectiveness: 100%
  - search.tool_choice.scope_awareness: 75%
  - search.retrieval.multi_hop: 67%
  - search.entity_lookup: 67%
  - search.retrieval.single_hop: 67%
  - search.retrieval.multi_hop_light: 67%
  - search.multi_hop_light: 67%
  - search.retrieval.precision: 67%
  - search.retrieval.recall: 67%
  - search.multi_hop: 60%
  - search.multihop: 60%
  - search.efficiency.tool_call_minimization: 50%
  - search.query_planning.multi-hop: 0%
  - search.strategy.source_prioritization: 0%
  - search.control.zero_result_guardrail: 0%
  - search.control.stage_adherence: 0%
  - search.control.answer_block_on_zero_results: 0%
  - search.control.stage_adherence.strict: 0%
  - search.tool_choice.code_for_computation_only: 0%
  - search.retrieval.robust_authoritative_search: 0%
  - search.strategy.query_formulation_refinement_and_verification: 0%
Strengths:
  - answered from recall despite zero search results
  - kept cost low with a single lightweight search attempt and no unnecessary tools
  - low-cost factual retrieval
Weaknesses:
  - no effective search query produced results
  - did not reformulate after zero results
  - issued a search call but returned no results

### Reasoning Models

#### reasoner-1 (gpt-5)
Overall: 45% success (10/22)
Skill scores:
  - code.math.date_calculation: 60%
  - code.math.arithmetic: 60%
  - code.data.aggregation: 60%
  - code.execution.iteration_budgeting: 8%
  - code.generation.codeblock_integrity: 0%
  - code.execution.token_economy: 0%
  - code.execution.syntax_self_check: 0%
  - code.stage_adherence: 0%
  - code.execution.strategy_handoff_on_failure: 0%
  - code.execution.offline_minimal_inline_structures: 0%
  - code.datetime.date_difference: 0%
Strengths:
  - kept code attempts to a single run (iteration budgeting)
  - lowest orchestration cost among peers
  - eventual handoff to answer after code provided no value
Weaknesses:
  - no executable code produced
  - ineffective search (no useful retrieval)
  - relied on non-code answer without tool-aligned computation/output

#### reasoner-2 (gpt-5-mini)
Overall: 62% success (13/21)
Skill scores:
  - code.math.multi_step: 100%
  - code.data.aggregation: 100%
  - code.date.arithmetic: 100%
  - code.execution.iteration_budgeting: 38%
  - code.execution.token_economy: 10%
  - code.generation.codeblock_integrity: 6%
  - code.execution.syntax_self_check: 0%
  - code.stage_adherence: 0%
  - code.execution.strategy_handoff_on_failure: 0%
  - code.execution.offline_minimal_inline_structures: 0%
  - code.datetime.date_difference: 0%
Strengths:
  - arrived at the correct concise numeric answer
  - iterated with multiple tool calls
  - kept Python syntactically valid (no execution errors)
Weaknesses:
  - inserted non-code/prose into a code block
  - no printed/output result from code executions
  - ineffective search (no useful retrieval)

#### reasoner-3 (Qwen/Qwen2.5-Coder-32B-Instruct)
Overall: 50% success (11/22)
Skill scores:
  - code.math.date_calculation: 100%
  - code.math.arithmetic: 100%
  - code.data.aggregation: 100%
  - code.data.tabular_parsing_ops_ranking: 100%
  - code.execution.strategy_handoff_on_failure: 50%
  - code.execution.iteration_budgeting: 31%
  - code.execution.token_economy: 30%
  - code.generation.codeblock_integrity: 0%
  - code.execution.syntax_self_check: 0%
  - code.stage_adherence: 0%
  - code.execution.offline_minimal_inline_structures: 0%
  - code.datetime.date_difference: 0%
Strengths:
  - produced the correct concise numeric answer
  - cost-efficient with a single code call and low token usage
  - date calculations
Weaknesses:
  - did not produce or display any code output (no executable/printed result)
  - overkill for simple factual lookups
  - produced no observable output from the code tool

### Answer Models

#### answer-1 (gpt-5)
Overall: 75% success (21/28)
Strengths:
  - returned a single exact number
  - used direct recall without retrieval
  - directly recalled the cultural fact
Weaknesses:
  - did not perform the required multi-hop reasoning
  - no verification / disambiguation checks
  - higher cost than answer-3

#### answer-2 (gpt-5-mini)
Overall: 87% success (13/15)
Strengths:
  - returned a single exact number
  - direct recall of well-known event
  - concise, exact output
Weaknesses:
  - ineffective query led to no results and no refinement attempt
  - defaulted to an unverified memorized answer
  - did not execute the multi-hop chain

#### answer-3 (meta-llama/Llama-3.3-70B-Instruct)
Overall: 62% success (8/13)
Strengths:
  - returned a single exact number
  - low overall overhead
  - immediate direct recall
Weaknesses:
  - performed a search that yielded no results and did not iterate/refine
  - failed to disambiguate role timing
  - did not verify the attribute definition

#### answer-4 (Qwen/Qwen3-32B)
Overall: 47% success (7/15)
Strengths:
  - returned a single exact number
  - correct direct recall
  - no tool usage
Weaknesses:
  - higher token usage compared to peers
  - no-hit search without query refinement or alternate strategy
  - no multi-hop reasoning or validation before finalizing

#### answer-math-1 (Qwen/Qwen2.5-Math-72B-Instruct)
Overall: 43% success (6/14)
Strengths:
  - returned a single exact number
  - correct direct recall beyond math domain
  - concise exact answer
Weaknesses:
  - misused the code tool by emitting non-Python content under the code stage
  - retrieved irrelevant/low-authority content and failed to process it
  - no error handling or validation pipeline

#### answer-math-2 (Qwen/Qwen2.5-Math-7B-Instruct)
Overall: 15% success (2/13)
Strengths:
  - attempted search before answering
  - produced the correct content via recall
  - concise single-entity output
Weaknesses:
  - produced an incorrect numeric answer
  - did not correctly align the two event dates before computing the difference
  - likely exact-match formatting or evaluator compliance issue
\end{lstlisting}

\medskip
\hrule
\medskip

\begin{lstlisting}[language={}, basicstyle=\ttfamily\scriptsize]
## Detailed Instructions

STEP 1 - REQUIRED
Based on the Problem and Context shown at the top, determine the next stage:
(search | code | answer).

- search: retrieve missing external information needed to answer the problem.
- code: write and execute Python code to compute or process data.
- answer: synthesize gathered information into the final answer.

STEP 2 - REQUIRED FORMAT
After deciding the next stage:

1. Reason inside <think>...</think>:
   - Why this stage is needed.
   - Which skills are required.
   - The relative importance of each skill.

2. Output the structured skill analysis inside <skill_analysis> tags:

<skill_analysis>
{
  "required_skills": [
    {"skill_id": "skill.id", "percentage": 50},
    {"skill_id": "skill.id", "percentage": 30}
  ],
  "reasoning": "Brief explanation of why these skills are needed."
}
</skill_analysis>

STEP 3
Choose the appropriate tool and model based on the skill analysis.

---

## Final Reminders

CRITICAL:
- The <skill_analysis> block is MANDATORY.
- It MUST appear BEFORE any tool call.

IMPORTANT:
- When calling a tool, you MUST specify the model parameter using the model alias (e.g., \"answer-1\", \"search-1\", \"reasoner-1\").
- Use exact model names from the Available Models section above.

Now, based on the Problem and Context shown at the top, analyze what should be the next stage (search|code|answer), provide the detailed skill analysis needed for the next stage in the <skill_analysis> tags and then choose an appropriate tool.
\end{lstlisting}

\end{tcolorbox}

\captionof{figure}{Agent orchestration instruction used for FRAMES, integrating the task query, execution history context, and the selected Skill Handbook to enable skill-aware orchestration.}
\label{fig:agent_orch_inst}

\end{document}